\documentclass[11pt]{article}

\usepackage[preprint]{acl}

\usepackage{times}
\usepackage{latexsym}
\usepackage{paralist}
\usepackage[T1]{fontenc}
\usepackage{subcaption}
\usepackage[utf8]{inputenc}

\usepackage{microtype}

\usepackage{inconsolata}

\usepackage{graphicx}

\usepackage{enumitem} 

\usepackage{enumitem}
\usepackage{booktabs}

\usepackage{multirow} 
\usepackage{amssymb}

\usepackage{makecell}
\usepackage{booktabs}

\newcommand\blfootnote[1]{%
  \begingroup
  \renewcommand\thefootnote{}\footnote{#1}%
  \addtocounter{footnote}{-1}%
  \endgroup
}

\newcommand\setalign{4pt}

\title{SEA-NLI: Natural Language Inference as a Lens into Southeast Asian Cultural Understanding}

\author{
Peerawat Chomphooyod\textsuperscript{1,*}, 
Jian Gang Ngui\textsuperscript{2},
Yosephine Susanto\textsuperscript{2},
\\
\textbf{Attapol T. Rutherford}\textsuperscript{1},
\textbf{Alham Fikri Aji}\textsuperscript{3},
\textbf{Sarana Nutanong}\textsuperscript{4},  
\\
\textbf{Can Udomcharoenchaikit}\textsuperscript{4,$\dagger$},  
\textbf{Peerat Limkonchotiwat}\textsuperscript{1,2,$\dagger$}\\
  \textsuperscript{1}Chulalongkorn University
  \textsuperscript{2}AI Singapore 
  \textsuperscript{3}MBZUAI
  \textsuperscript{4}VISTEC,
   \\
  \texttt{canu\_pro@vistec.ac.th}, \texttt{peerat@aisingapore.org}
  \\ 
  \href{https://huggingface.co/datasets/aisingapore/SEA-NLI}{\includegraphics[height=1em]{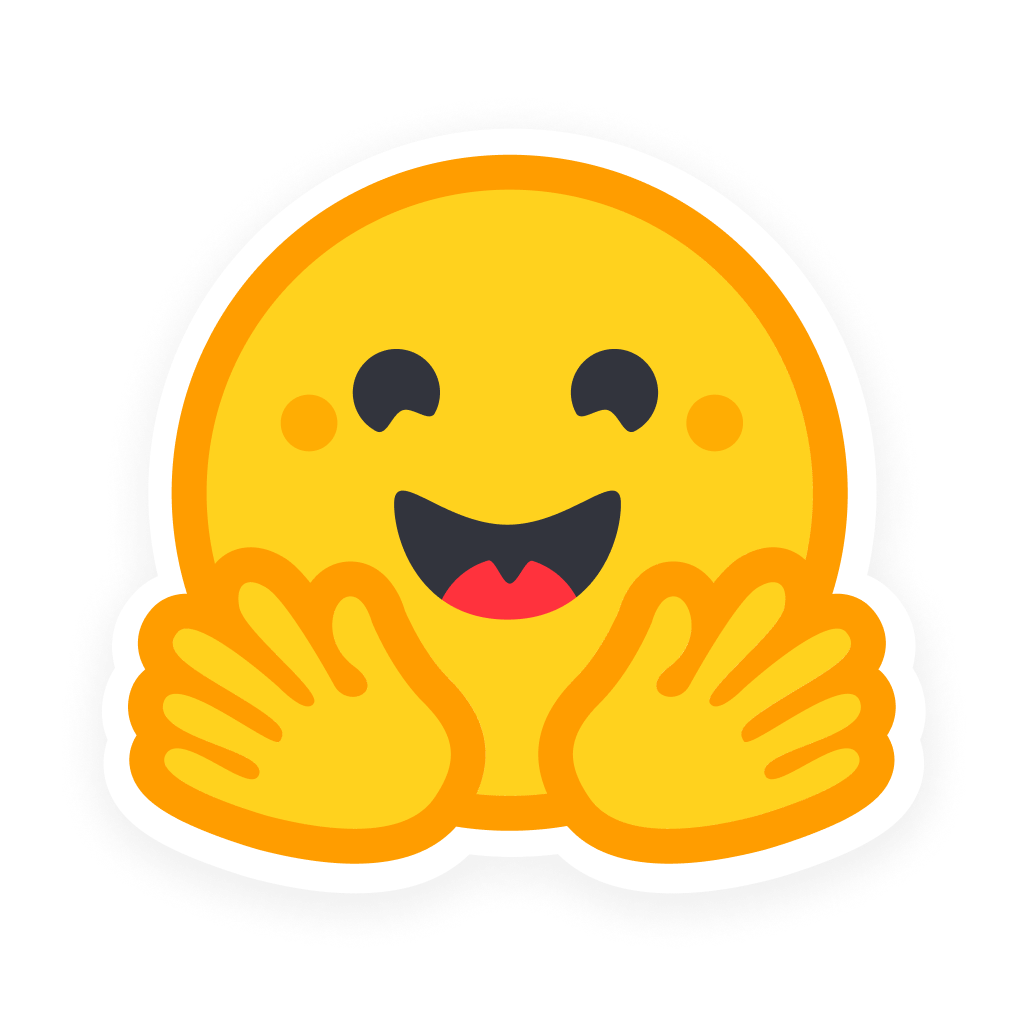}\ aisingapore/SEA-NLI}
  }

\begin{document}
\maketitle

\blfootnote{\textsuperscript{*}Work conducted during Research Internship at AI Singapore}
\blfootnote{\textsuperscript{$\dagger$}Corresponding authors}

\begin{abstract}
Frontier LLMs perform well in Western contexts, but remain poorly tested on underrepresented cultures such as those in Southeast Asia (SEA).
Existing NLI benchmarks are largely Western-centric, translation-derived, or monolingual, limiting their ability to measure culturally grounded reasoning.
We introduce \textbf{SEA-NLI}, a native, culturally grounded NLI benchmark covering eight SEA countries in English and native regional languages, verified by native speakers.
Across 17 encoder and decoder models, we observe a low performance from all models, especially for knowledge-intensive categories such as Languages and Science and Technology.
Our analysis shows that failure cases mainly stem from missing SEA cultural knowledge: SEA-adapted models and culture-aware prompting improve performance, while CoT prompting offers limited gains.
\end{abstract}
\section{Introduction}

Frontier large language models (LLMs) excel in Western settings but often underperform in underrepresented, data-scarce cultures, largely due to training data skewed toward dominant Western linguistic and cultural paradigms.
Culturally grounded evaluations---including cross-cultural reasoning~\cite{romero2024cvqaculturallydiversemultilingualvisual,satar-etal-2025-seeing,lin2026culturallbenchmarkingmultilingualmulticultural,kabir2026xcrbenchmultitaskbenchmarkevaluating} and local knowledge retrieval~\cite{bogdanova2026flanssemeval2026task7,li2026raveneabenchmarkmultimodalretrievalaugmented}---are essential for surfacing these biases and improving cultural equity.
Prior work also uses natural language inference (NLI) to probe cultural knowledge~\cite{mahendra-etal-2021-indonli,huang-yang-2023-culturally}. 
In cultural NLI, the correct label can hinge on culturally situated facts; for instance, resolving the following NLI query requires knowing that \textit{Golden Pillow} is a well-known Thai durian cultivar rather than an actual pillow:
\begin{list}{}{
  \setlength{\leftmargin}{1em}
  \setlength{\rightmargin}{1em}
}
\item \noindent\textbf{Premise:} Dan bought a \textit{Golden Pillow} from a Thai fruit stall.\\
\textbf{Hypothesis:} Dan bought a bedding item.\\
\textbf{Gold label:} \textsc{contradiction}.
\end{list}

As summarized in Table~\ref{tab:dataset_comparison_v2}, existing NLI datasets are predominantly Western-centric, translation-derived, and monolingual, with none combining native multilingual SEA coverage and cultural knowledge.
Multilingual datasets are often built by translating English texts~\cite{conneau-etal-2018-xnli,ham-etal-2020-kornli,heredia-etal-2024-xnlieu,Htet2025}, which can introduce errors~\cite{agrawal-etal-2024-translation} and obscure local meanings, idioms, socio-cultural norms, and value systems~\cite{singh-etal-2025-global,susanto2025sea}.
As a result, they often retain Western-centric source perspectives, leaving a gap in resources that faithfully represent Southeast Asian (SEA) cultures.
This gap is critical because SEA accounts for 10\% of the world's population, with over 700 million people,\footnote{\url{https://population.un.org/wpp/}} yet remains underrepresented in cultural understanding research.

\begin{table*}[t!]
\centering
\scalebox{0.68}{
\begin{tabular}{@{}lllclll@{}}
\toprule
\textbf{Dataset} & \textbf{Test Size} & \textbf{Lang} & \textbf{Culture} & \textbf{Data Source} & \textbf{Key Contribution} \\ 
\midrule
\multicolumn{6}{l}{\textit{\textbf{English Benchmarks}}} \\
SNLI \cite{bowman-etal-2015-large} & 10k & EN & Western & Image captions (Flickr30k) & Pioneering large-scale NLI benchmark \\
MNLI \cite{williams-etal-2018-broad} & 20k & EN & Western & 10 Genres (Letters, Fiction, etc.) & Multi-genre cross-domain evaluation \\
\midrule
\multicolumn{6}{l}{\textit{\textbf{Multilingual \& SEA Benchmarks}}} \\
XNLI \cite{conneau-etal-2018-xnli} & 75k & 15 & Western & Translation (MNLI) & Cross-lingual consistency benchmark \\ 
IndoNLI \cite{mahendra-etal-2021-indonli} & 5.2k & IDN & Indo. & Wikipedia, News, and Web & Challenging native-sourced IDN NLI \\
ViNLI \cite{huynh-etal-2022-vinli} & 3k & VNM & Viet. & Online news articles & High-quality VNM NLI \\
Myanmar XNLI \cite{Htet2025} & 5k & MYA & Western & Translation (XNLI) & Burmese low-resource evaluation \\
\midrule
\multicolumn{6}{l}{\textit{\textbf{Cultural Benchmarks}}} \\
CALI \cite{huang-yang-2023-culturally} & 2.7k & EN & US/India & NormBank & Cultural label disagreement study \\
\textbf{SEA-NLI (ours)} & \textbf{2.1k} & \textbf{SEA} & \textbf{SEA} & \textbf{SEA Wikipedia} & \textbf{SEA cultural knowledge evaluation} \\
\bottomrule
\bottomrule
\end{tabular}}
\vspace{-3mm}
\caption{NLI dataset comparison. SEA-NLI uniquely combines SEA cultural grounding with multilingual coverage.}
\vspace{-5mm}
\label{tab:dataset_comparison_v2}
\end{table*}

Therefore, this paper addresses three main research questions on the SEA NLI dataset construction and evaluation:
\begin{compactitem}[\hspace{\setalign}•]
    \item \textbf{RQ1:} How to build a robust SEA NLI dataset that accurately reflects SEA cultures?
    
    \item \textbf{RQ2:} How does LLM performance change when evaluated on SEA cultures rather than Western ones?

    \item \textbf{RQ3:} What gaps remain in current models, and how can they be mitigated?
\end{compactitem}

To answer \textbf{RQ1}, we propose \textbf{SEA-NLI}, a culturally-grounded NLI dataset for Southeast Asia.
SEA-NLI covers cultural topics from eight SEA countries and languages (\texttt{COUNTRY\_NAME:LANGUAGE}): {Cambodia:KHM}, {Myanmar:MYA}, {Malaysia:ZSM}, {Thailand:THA}, {Singapore:TAM}, {Philippines:FIL}, {Indonesia:IND}, and {Vietnam:VIE}, with texts in both English and SEA languages.
By doing so, SEA-NLI also serves as a diagnostic tool: correct English but incorrect SEA suggests a language-understanding gap, failure in both suggests missing cultural knowledge, and correct SEA but incorrect English suggests cross-lingual misalignment.

We construct SEA-NLI via LLM generation with human verification to ensure culturally grounded, high-quality examples.
Annotators first select cultural concepts from Wikipedia and draft candidate pairs while mitigating common artifacts (e.g., negation matching, length bias, and noun-phrase variation).
We then apply lexical and semantic filters to remove easy cases and regenerate samples until these artifacts are eliminated.
Native annotators from all eight countries review and revise each example for cultural accuracy and relevance.
Finally, we split SEA-NLI into \textbf{normal} (reduced premise--hypothesis overlap) and \textbf{hard} (regenerated pairs with no cultural topic overlap).

We further evaluate current models on SEA-NLI (\textbf{RQ2} and \textbf{RQ3}) through 5 experiments on 17 models.
For \textbf{RQ2}, SEA-NLI is challenging: average performance drops by 11.60\%/13.17\% on the hard set in SEA/English compared to the normal set, with low scores on knowledge-intensive categories such as Languages and Science and Technology.
For \textbf{RQ3}, model failure cases are driven mainly by missing SEA-specific cultural knowledge rather than generic reasoning limits: SEA-adapted models and culture-aware prompting improve performance, while CoT provides limited benefit.
These results show that improving SEA-NLI performance requires SEA-specific adaptation, rather than relying only on high-quality English data or model reasoning ability.

Our contributions are as follows. 
\begin{compactitem}[\hspace{\setalign}•]
    \item We release SEA-NLI, a culturally-grounded NLI dataset for Southeast Asia, reviewed, corrected, and localized by native speakers across eight countries and languages.

    \item We establish a benchmark for evaluating cultural understanding in SEA and show that both encoder and decoder models experience substantial performance drops on culturally challenging examples, especially in our hard subset.
    
    \item By providing each example in a SEA language and English, we enable diagnostic evaluation that disentangles language-understanding gaps, missing cultural knowledge, and cross-lingual misalignment; we further identify failure modes in cultural reasoning, showing that culturally aware prompting improves LLM performance.
\end{compactitem}

\section{Related Works}

\subsection{Natural Language Inference (NLI)}
NLI determines the logical relationship between a premise and a hypothesis.
Given the premise as true, a model classifies the pair as entailment (the hypothesis is also true), contradiction (the hypothesis is inconsistent with the premise), or neutral (whether the hypothesis is true or not cannot be determined given the premise).
%
%
Popular NLI datasets, such as SNLI~\cite{bowman-etal-2015-large}, MNLI~\cite{williams-etal-2018-broad}, and SICK~\cite{marelli-etal-2014-sick}, are largely English-centric, reflecting Anglosphere linguistic conventions and cultural perspectives.
Multilingual extensions, most notably XNLI~\cite{conneau-etal-2018-xnli}, translate English texts into other languages.
This translation-based paradigm remains common in later language-specific, multilingual, and domain-specific benchmarks~\cite{wijnholds-moortgat-2021-sick,yanaka-mineshima-2022-compositional,ebrahimi-etal-2022-americasnli,aggarwal-etal-2022-indicxnli,heredia-etal-2024-xnlieu,obadic-etal-2023-c,Htet2025,Ogul2025-gn}.
While expanding linguistic coverage, these datasets often inherit cultural assumptions from English source texts rather than representing local contexts.

To reduce reliance on translated English data, native-source NLI datasets have been developed for non-English languages~\cite{hu-etal-2020-ocnli,mahendra-etal-2021-indonli,huynh-etal-2022-vinli,VANHUYNH2026130109}.
These resources address language-specific data scarcity and better reflect local usage, but do not systematically evaluate cultural knowledge.
Beyond data sources, CALI~\cite{huang-yang-2023-culturally} studies how cultural background affects NLI label disagreement across annotator groups~\cite{plank-2022-problem}, but is limited to English and two cultures.
Thus, despite progress in multilingual and native-source NLI, no comprehensive dataset covers major Southeast Asian languages with an explicit focus on SEA cultural and regional knowledge.

\subsection{NLI Synthetic Data Generation}

Human annotation remains reliable for producing correct NLI examples, but scaling diverse and challenging examples is difficult~\cite{liu-etal-2022-wanli}.
LLMs enable more controlled data generation, especially when combined with filtering or selection strategies to improve quality.
Prior work shows that synthetic data targeting domain gaps~\cite{hosseini-etal-2024-synthetic}, spurious correlations~\cite{wu-etal-2022-generating}, and complex examples for fine-tuning or distillation~\citep{stacey-rei-2024-distilling,stacey-etal-2026-improving} can improve robustness and generalization.
Human-AI collaborative frameworks, such as WANLI~\cite{liu-etal-2022-wanli} and CALI~\cite{huang-yang-2023-culturally}, further improve synthetic data by combining LLM generation with human oversight, with CALI explicitly eliciting cultural knowledge during annotation.
Together, these studies suggest that strategically curated, complex, and culturally informed synthetic data can reduce dataset biases and support more reliable NLI evaluation.
\section{SEA-NLI}

\begin{figure}[h!]
  \centering
  \includegraphics[width=0.9\linewidth]{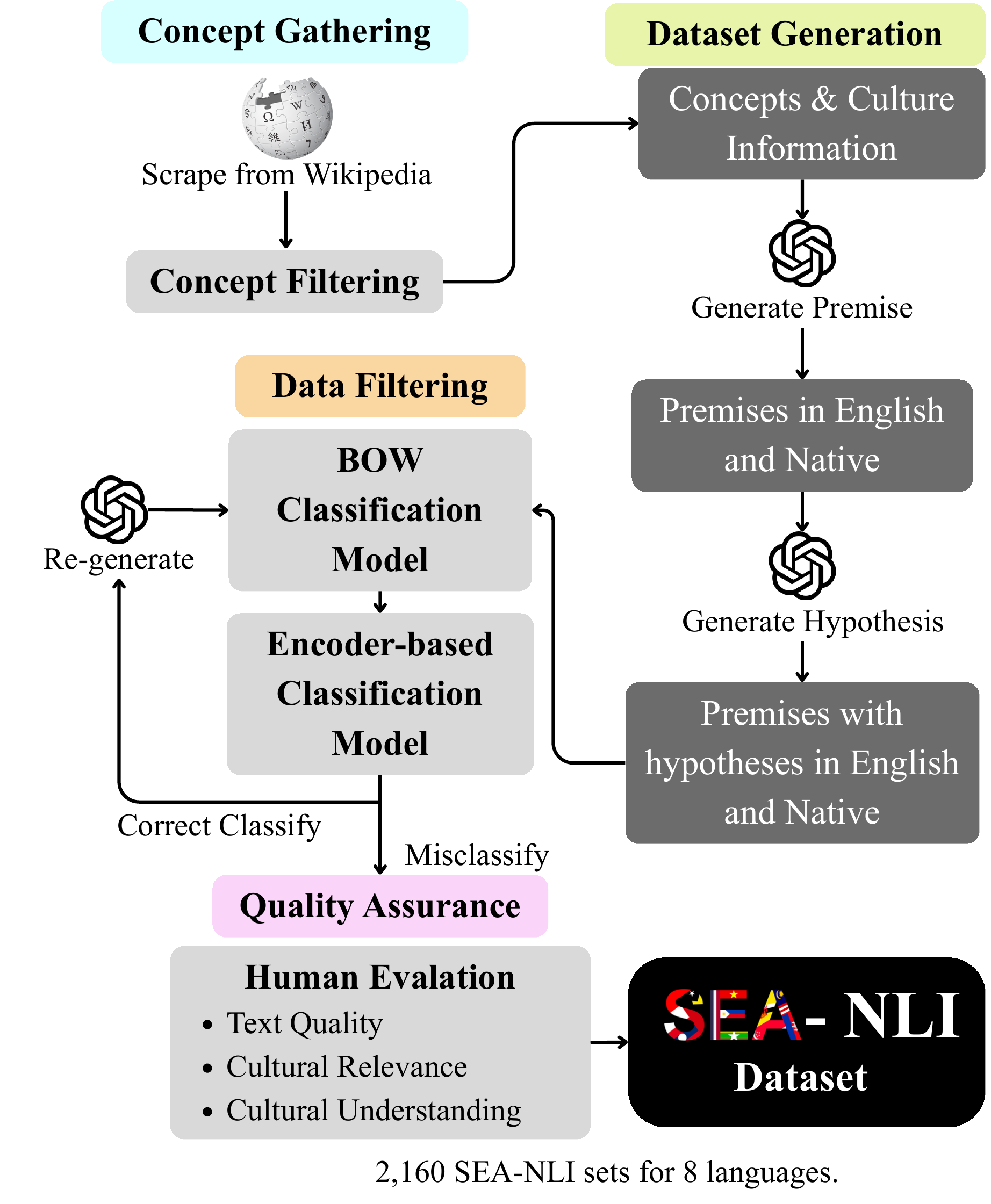}
  \vspace{-3mm}
  \caption{The complete process for creating SEA-NLI.}
  \vspace{-5mm}
  \label{fig:SEA-NLI-frist-img}
  \vspace{-0.5em}
\end{figure}

The creation of the SEA-NLI dataset follows four key steps, as shown in Figure \ref{fig:SEA-NLI-frist-img}.
We describe each step as follows.
    
    
    

\subsection{Concept Gathering} \label{subsec:concept_garthering}

Following the cultural taxonomy of prior works~\cite{zhang-etal-2025-culturesynth,alkhamissi-etal-2026-hire}, we construct SEA-NLI to evaluate \emph{culture-as-knowledge}, i.e., whether models can apply cultural knowledge to infer the correct NLI label.
As shown in Figure~\ref{fig:SEA-NLI-frist-img}, we collect 7,904 Wikipedia topics across 10 categories: (i) landmarks, (ii) clothing, (iii) musical instruments, (iv) languages, (v) crime, (vi) science and technology, (vii) cuisine, (viii) education, (ix) politics, and (x) economy, with annotators pre-screening unrelated topics for each country.
In addition, we discuss the annotator guidelines in Appendix~\ref{appendix:screening}.
For each topic in Wikipedia, we extract its {title}, {category}, and {summary}: the page title, category hierarchy, and introductory paragraph, respectively.
We used Wikipedia pages and topics from 2024-2026. 

\begin{figure*}[h!]
  \centering
  \includegraphics[width=0.95\linewidth]{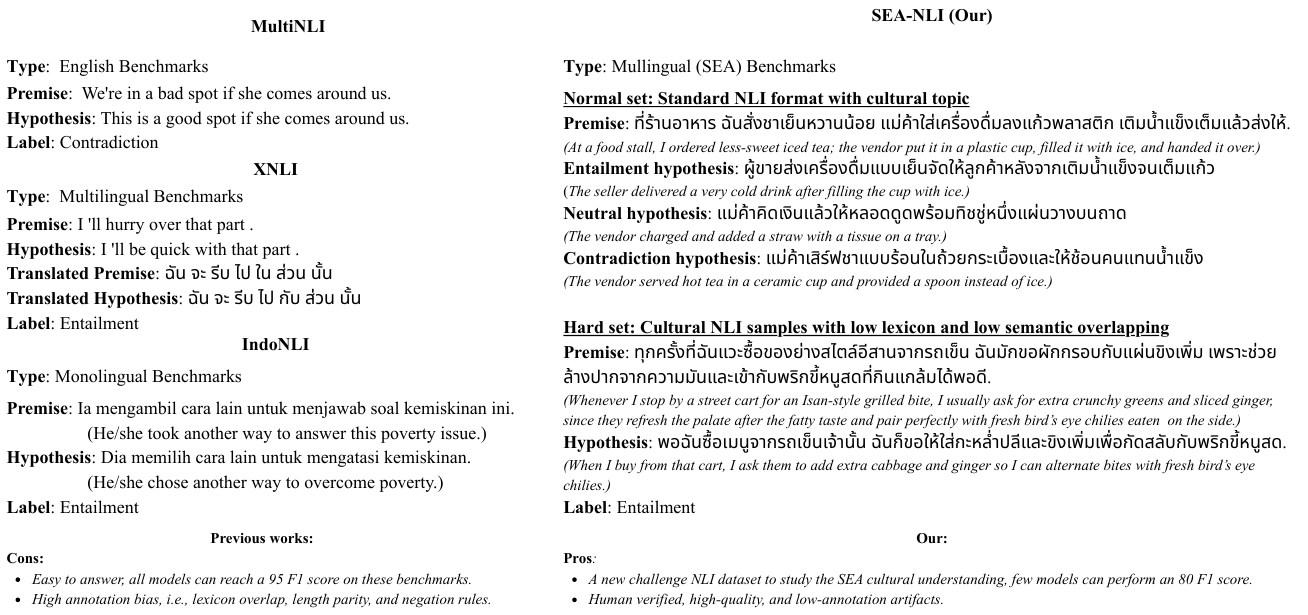}
  \vspace{-3mm}
  \caption{Comparison of SEA-NLI with the existing NLI datasets}
  \vspace{-7mm}
  \label{fig:compare-dataset}
\end{figure*}

\subsection{Dataset Generation} \label{subsec:data_generation}

We use GPT-5.2 to generate SEA-NLI through three stages: (i) premise generation, (ii) hypothesis generation, and (iii) quality improvement.
All prompts and examples are provided in Appendix~\ref{apd:dataset_generation_prompts} and \ref{subsec:regeneration_prompt}.

\noindent
\textbf{Premise Generation.}
Given a topic's title, category, and summary from Section~\ref{subsec:concept_garthering}, GPT-5.2 generates premises in both English and the topic's native language.
According to previous works~\cite{alkhamissi-etal-2024-investigating, wang-etal-2024-countries}, the system prompt includes the following constraining instructions, to improve cultural accuracy and generation quality:

\begin{compactitem}[\hspace{\setalign}•]
    \item \textbf{Role-prompting and persona adoption:} Assigning a professional identity and cultural background to ensure culturally relevant premises.
    \item \textbf{Constraint-based generation:} Enforcing strict rules regarding sentence length and the inclusion of localized keywords.
    \item \textbf{Concept adaptation:} Allowing the LLM to rewrite concept titles for each country, e.g., \href{https://en.wikipedia.org/wiki/Shrimp_curry}{shrimp curry} to ``Phuket shrimp curry'', which is clearer for Thai native speakers. If the original title is already specific, it remains unchanged.
    \item \textbf{Concept summary:} Since Wikipedia summaries may include information irrelevant to the target culture, e.g., ``\href{https://en.wikipedia.org/wiki/Shrimp_curry}{Phuket shrimp curry}'' may include Burmese or Indonesian variants, we let the LLM select only paragraphs related to the target concept.
\end{compactitem}

\noindent
\textbf{Hypothesis Generation.}
Given each generated premise, GPT-5.2 generates hypotheses for the three NLI labels: \texttt{entailment}, \texttt{neutral}, and \texttt{contradiction}.
As in premise generation, we apply constraint-based prompting to avoid trivial samples solvable through superficial linguistic patterns.
Furthermore, to mitigate annotation artifacts~\cite{zhou-bansal-2020-towards,mccoy-etal-2019-right,belinkov-etal-2019-adversarial}, we enforce the following rules:

\begin{compactitem}[\hspace{\setalign}•]
    \item \textbf{Length parity:} To prevent length-based bias, the word count difference between any two hypotheses for the same premise is strictly limited to a maximum of four words.
    \item \textbf{Lexical overlap mitigation:} The prompt explicitly forbids high word overlap between the premise and the entailment hypothesis, a common issue in standard NLI datasets.
    \item \textbf{Anti-negation rules:} Contradictions cannot be formed by simple negations (e.g., merely adding ``not'' or ``no'').
    \item \textbf{Noun phrase variation:} The prompt prevents the reliance on generic-to-specific noun phrase substitutions (e.g., replacing ``animal'' with ``dog'') to trivially generate entailments.
    \item \textbf{Hedging and intensity bans:} For neutral hypotheses, unverifiable hedging words (e.g., ``might,'' ``possibly'') and intensity cues are strictly prohibited.
    \item \textbf{Avoidance of superlatives:} Absolute quantifiers (e.g., ``all,'' ``none,'' ``always,'' ``only'') are banned in neutral and contradiction labels, as models frequently exploit these statistical cues to predict non-entailment.
\end{compactitem}

\noindent
\textbf{Quality Improvement.}
In a pilot study, we found that frequent words were spuriously associated with specific NLI labels, enabling shortcut-based prediction.
This problem also occurred in previous NLI datasets~\cite{bowman-etal-2015-large,mahendra-etal-2021-indonli}.
To mitigate this problem, we banned these terms in the final generation prompt and used GPT-5.2 to rewrite premises that explicitly explained cultural concepts.
These steps reduce lexical and contextual shortcuts, requiring models to rely more on intrinsic cultural knowledge for NLI classification.
In addition, we reuse the adapted concept and summary from the premise generation step to improve cultural grounding.
Without rewriting, output quality was substantially less reliable, as shown in Appendix~\ref{apd:dataset_improvement_prompts}.

\subsection{Data Filtering}
\label{subsec:data_filtering}
Despite our best efforts to remove annotation artifacts, there will still be some left in the dataset, including word matching, sentence length, negation, etc. 
These artifacts may allow for models to try to find shortcuts for prediction~\cite{zhou-bansal-2020-towards,mccoy-etal-2019-right,belinkov-etal-2019-adversarial}.
To remove these problems, we propose a data filtering system using lexical and semantic filters to identify \emph{easy examples} for all premise-hypothesis pairs.

\noindent
\textbf{Lexicon Filtering.}
We extract the lexical features using a Bag-of-Words (BoW) from the English examples of our test set, and train a logistic regression (LR) using BoW as the feature to predict the class of premise-hypothesis pairs\footnote{We cannot apply this to SEA languages because no robust word tokenizer covers all of them.}.
We remove samples that the LR model can correctly predict since these examples are trivially solvable with lexical matching without cultural knowledge requirements.

\noindent
\textbf{Semantic Filtering.}
We further filter easy samples using DeBERTa-v3~\cite{he2023debertav3improvingdebertausing}, a strong NLI encoder model provided by \citet{Laurer_van_Atteveldt_Casas_Welbers_2024}.
Because DeBERTa-v3 lacks SEA cultural knowledge, correctly predicted samples are treated as shortcut-solvable and regenerated until both LR and DeBERTa-v3 misclassify them.

\noindent
\textbf{Regenerate Samples.}
We regenerate samples that fail either lexical or semantic filtering.
The regeneration prompt uses stricter premise-hypothesis constraints to reduce annotation bias (Appendix~\ref{apd:dataset_improvement_prompts}).
Regenerated samples are then filtered again with the same systems.
We repeat this for four rounds: only 2.07\% (492/23,712) pass in the first round, with later rounds adding 1.23\%, 1.28\%, and 0.82\%; we stop after four rounds due to diminishing returns.

\noindent
\textbf{Sample Splitting.}
As shown in Figure~\ref{fig:compare-dataset}, we split SEA-NLI into \emph{normal} and \emph{hard} sets.
The \emph{normal} set contains samples that pass the first filtering round without regeneration, with each premise paired with three hypotheses to follow the standard NLI format.
The \emph{hard} set contains regenerated premise-hypothesis pairs, making it imbalanced but more culturally demanding.
Together, the \emph{normal} and \emph{hard} sets test whether models can move beyond lexical or semantic shortcuts as crutches and rely on deeper SEA cultural understanding.

\subsection{Quality Assurance} \label{subsec:qa}

To ensure all samples met our quality threshold, we recruited 27 annotators from 8 countries to validate the dataset.
We describe the annotator metrics and process below.
Full annotator details and guidelines are provided in Appendix~\ref{apd:annotator_details}.

\noindent
\textbf{Task Metrics.}
Annotators checked four aspects: (i) label correctness, (ii) cultural relevance of each topic (Likert scale 1-5), (iii) cultural understanding required for each review topic (Likert scale 1-5), and (iv) issue flags (see Appendix~\ref{apd:human_evaluation_metrics}).
We use a majority vote for the final label and average scores within each annotator group.

\noindent
\textbf{Task Guidelines.}
Annotators reviewed SEA-NLI in three steps.
First, they evaluated GPT-5.2 output generated using pilot prompt techniques from Section~\ref{subsec:data_generation} (120 samples per country) to validate baseline translation, labels, and cultural relevance (see Appendix~\ref{apd:pilot_generation_prompts}).
Second, we verified labels and LLM outputs by reviewing examples generated without filtering or refinement.
%
Third, they reviewed the final SEA-NLI to ensure that regenerated samples are high-quality and require stronger cultural knowledge.
%
We expected that agreement and cultural-relevance scores of the \emph{second} and \emph{third} tasks from annotators would be better than those of the first task.

\noindent
\textbf{Results from Annotators.}
Annotator agreement is consistently high across the three steps (0.97, 0.99, and 0.96), while cultural relevance increases from 4.15 to 4.36 and 4.47.
This highlights that our datasets are more culturally relevant than vanilla GPT-5.2-generated texts.
Also, the regenerated samples did not damage any semantic or quality of data, as we got higher quality scores from annotators compared to samples before the regenerated process (the third task). 
We also find that 8.89\% of labels were changed and 1.48\% of samples were flagged for quality issues.
These results show that GPT-5.2 can generate reliable SEA-NLI candidates when properly prompted, but human validation remains necessary to ensure label correctness, quality, and cultural relevance.
All the scores and how we calculate them are discussed in Appendix~\ref{apd:annotation_agreement}. 

\begin{figure*}[h!]
  \centering
  \vspace{-3mm}
  \includegraphics[width=0.95\linewidth]{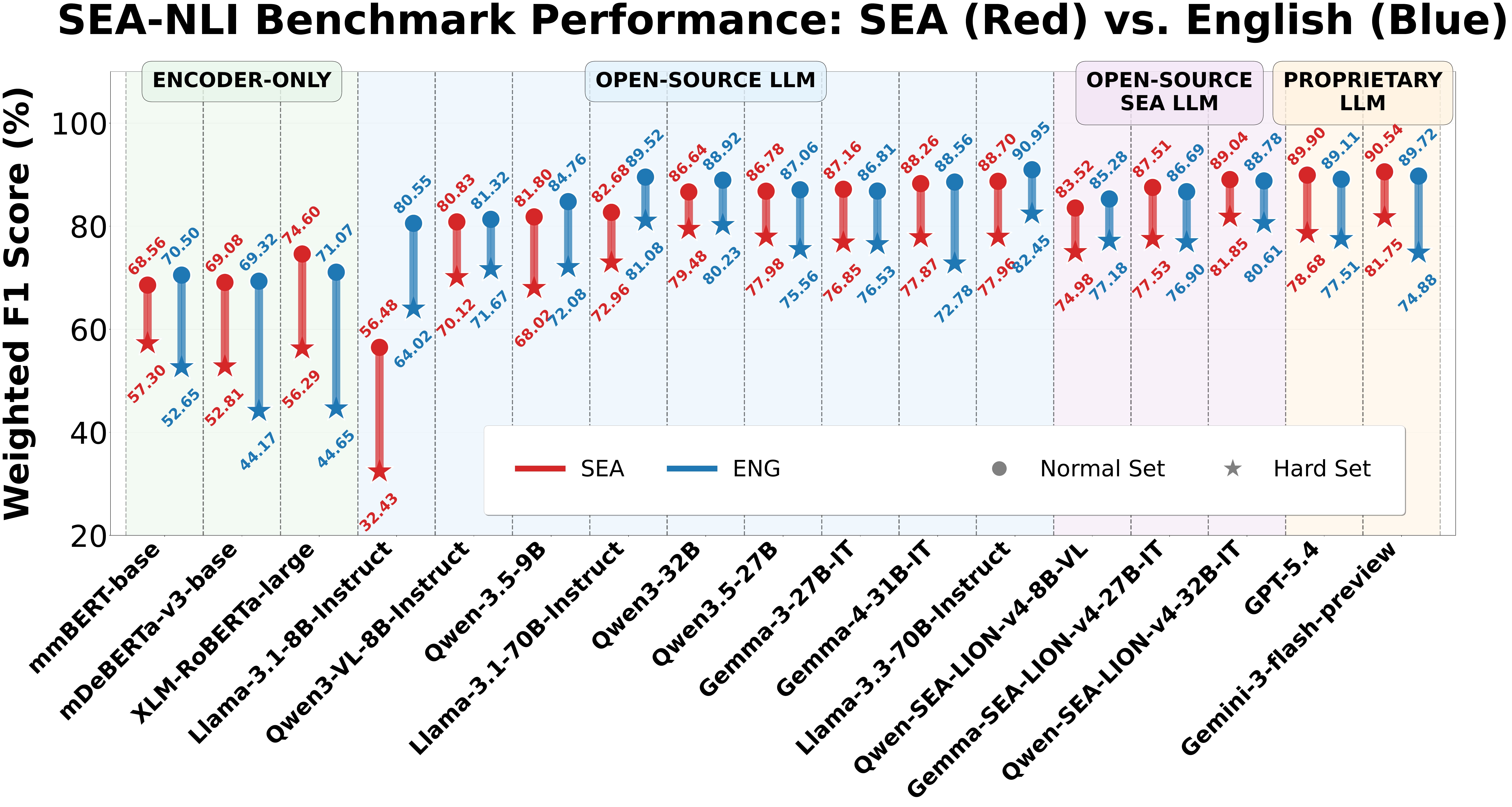}
  \vspace{-5mm}
  \caption{The overall performance of 17 models on SEA-NLI, where we compare the performance of normal and hard sets on SEA and English languages.}
  \vspace{-5mm}
  \label{fig:main_results}
\end{figure*}

\subsection{SEA-NLI Summary}
\begin{figure}[h!]
  \centering
  \vspace{-2mm}
  \includegraphics[width=1\linewidth]{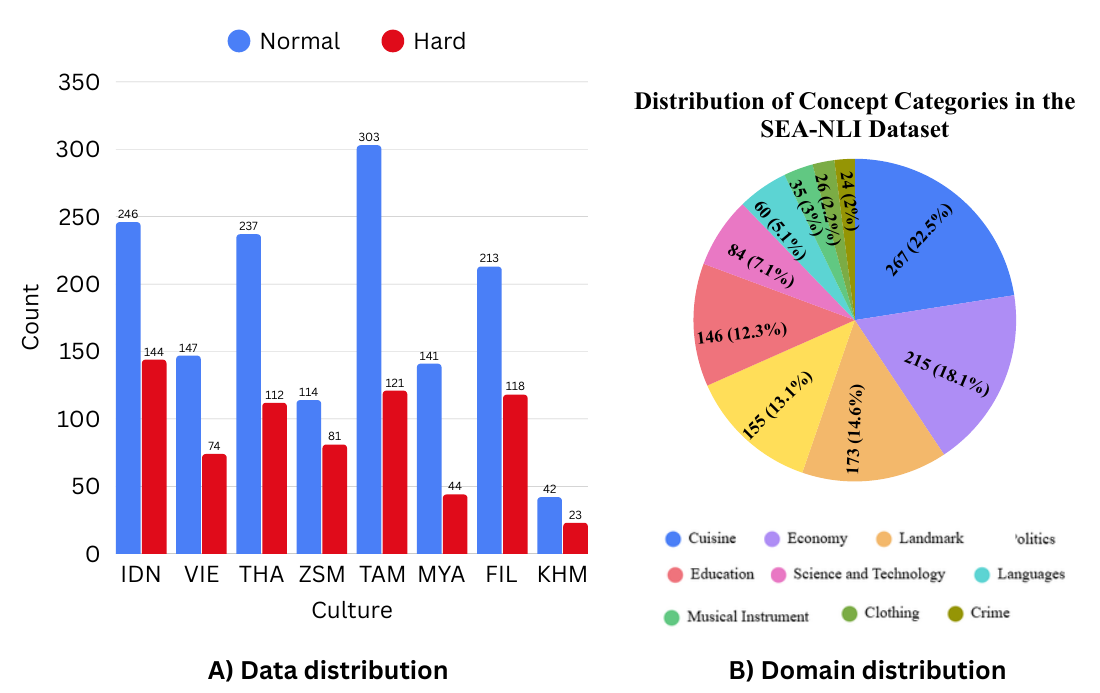}
  \vspace{-8mm}
  \caption{Data statistics of SEA-NLI.}
  \vspace{-8mm}
  \label{fig:statistic}
\end{figure}

We summarize the dataset statistics and distributions in Figure~\ref{fig:statistic}.
SEA-NLI contains 1,443 \emph{normal} and 717 \emph{hard} samples across 10 cultural categories, 8 countries, and 8 SEA languages.
%
SEA-NLI is the first NLI dataset focused on Southeast Asian cultural knowledge.
We ensure dataset quality through full annotator review of labels, cultural relevance, and translations, with annotator results confirming that the samples are culturally relevant and of high quality.
Each sample includes both the target culture's native language and an annotator-verified English translation from GPT-5.2, enabling analysis of SEA-English inconsistency.
Full statistics are provided in Appendix~\ref{apd:dataset_statistic}.
In addition, we conduct a bias study in SEA-NLI in Appendix~\ref{appendix:bias}.

\section{Experimental Setup}

\noindent
\textbf{Models.}
To answer \textbf{RQ2} and \textbf{RQ3}, we employ 17 models across encoder-based and decoder-based models to be evaluated on the SEA-NLI benchmark.
For the encoder-based evaluation, we fine-tune the pre-trained base models: XLM-R~\cite{conneau-etal-2020-unsupervised}, mmBERT~\cite{marone2025mmbertmodernmultilingualencoder}, and mDeBERTa~\cite{he2023debertav3improvingdebertausing}, using SNLI~\cite{bowman-etal-2015-large} and XNLI~\cite{conneau-etal-2018-xnli} datasets. 
%
%
For the decoder-based architectures, we evaluate the models in a zero-shot setting utilizing Qwen-3.5~\cite{yang2025qwen3technicalreport}, Llama-3.1/3.3~\cite{grattafiori2024llama3herdmodels}, and the SEA regionally specialized LLMs: SEA-LION-Qwen-v4 and Gemma-SEA-LION-v4~\cite{ng-etal-2025-sea}, where the prompt is detailed in Figure~\ref{fig:base_prompt}a. 
Additionally, we evaluate leading proprietary models via API, namely GPT-5.4 and Gemini-3-Flash. 
%

\noindent
\textbf{Metrics.}
We evaluate the NLI classification performance using the weighted-averaged F1-score, computed on both SEA languages and English-translated versions of the dataset separately. 
%


\section{Experimental Results}

\subsection{Main Results}
\label{subsec:main_result}

To answer \textbf{RQ2}, we summarize results from 17 models on SEA-NLI normal and hard sets in Figure~\ref{fig:main_results}.
Full results are reported in Table~\ref{tab:evaluation_result_normal}.

\noindent
\textbf{Normal vs. Hard Sets.}
The hard set is substantially more difficult than the normal set for all model families.
Averaged across all models, performance drops from 81.89/84.05 on the normal set to 70.29/70.88 (11.60\%/13.17\% drop) on the hard set in SEA/English.
%
This degradation is especially pronounced for encoder-based models, indicating that the hard set removes many shortcut-solvable examples and requires stronger reasoning beyond superficial lexicon or generic semantic cues.

\noindent
\textbf{English vs. SEA Languages.}
Cross-lingual robustness remains inconsistent.
On the normal set, many models perform slightly better in English, but this pattern is less stable on the hard set, where the SEA--English gap widens.
For example, \texttt{Llama-3.1-8B-Instruct} and other Llama models show much larger SEA--English gaps than other models.
This suggests that strong English NLI performance does not necessarily transfer to SEA languages in culturally challenging settings.

\noindent
\textbf{Effect of SEA Adaptation.}
As shown in Figure~\ref{fig:main_results}, we also observe that SEA-adapted models generally improve over their corresponding base models, especially on SEA-language evaluation.
For example, \texttt{Qwen-SEA-LION-v4-8B-VL} outperforms \texttt{Qwen3-VL-8B-Instruct} by 2.69/3.96 points on the normal set and 4.86/5.51 points on the hard set.
These results emphasize that both SEA-specific adaptation and model scaling are important for robust culturally grounded NLI.

\noindent
\textbf{Summary.}
The hard set introduces a new NLI challenge that is less solvable by lexical overlap, sentence patterns, or generic NLI reasoning alone.
Because these samples survive both the lexical and semantic filters, models must rely more heavily on culturally-grounded knowledge to infer the presence of an entailment relation correctly.
The broad performance drop across all models, including frontier models, highlights cultural understanding as a key remaining challenge in SEA-NLI.

\subsection{Category Results} \label{subsec:cat_results}
\begin{figure}[h!]
\vspace{-2mm}
  \centering
   \includegraphics[width=1\columnwidth, clip]{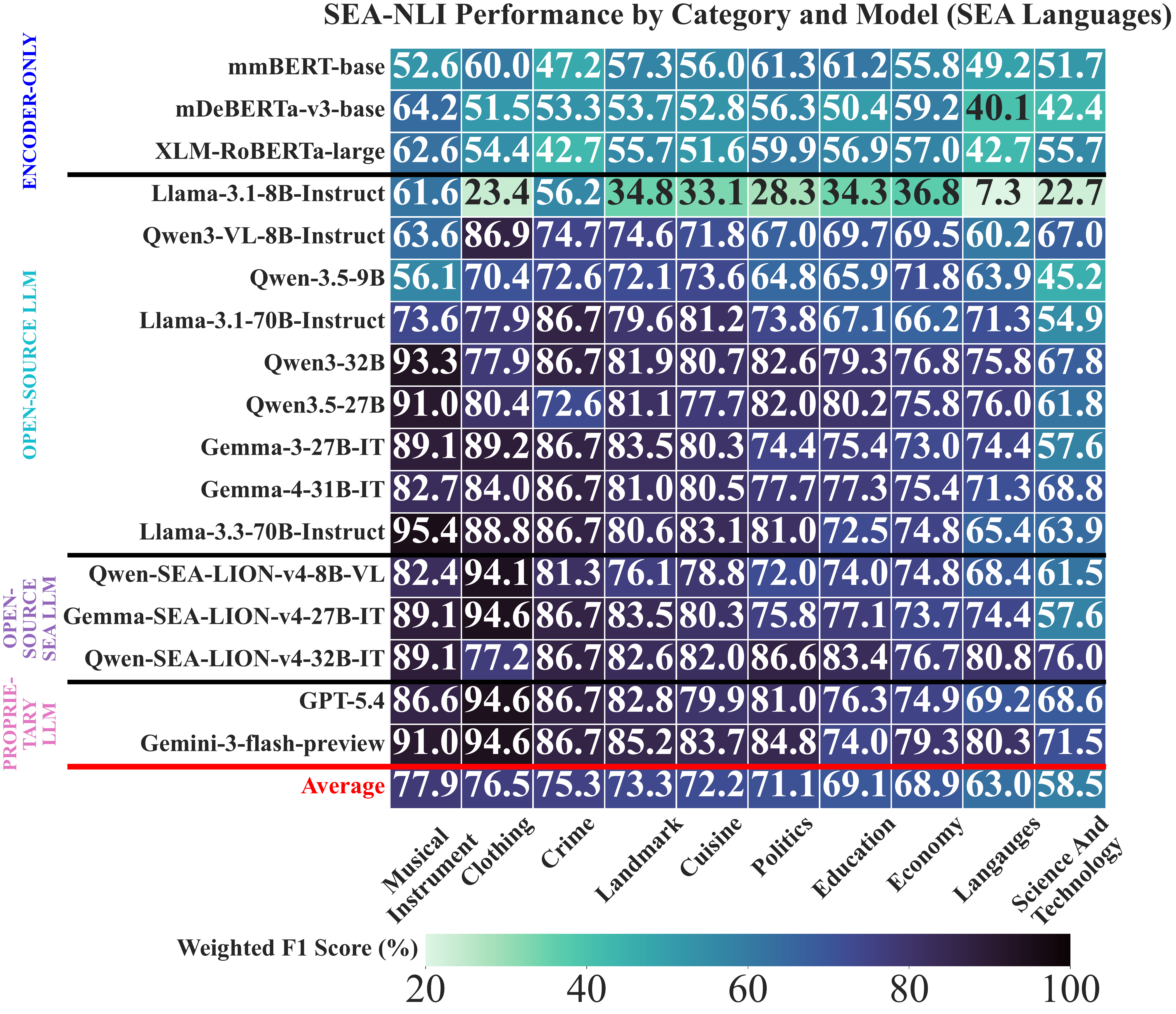}
  \vspace{-8mm}
  \caption{Weighted F1 performance across cultural concept categories on SEA languages (Hard set). We mixed all languages within the same category.}
  \vspace{-5mm}
  \label{fig:accuracy_on_category_native_vs_lang}
\end{figure}

We answer \textbf{RQ2} by evaluating the performance of models using the category results.
As shown in Figure~\ref{fig:accuracy_on_category_native_vs_lang}, model performance varies substantially across cultural concept categories.
Overall, models achieve the strongest results on more visually or commonly represented cultural topics, such as Musical Instrument, Cuisine, and Landmark, with average F1 scores around 72\%-78\% across the dataset.
In contrast, categories that require more fine-grained cultural or contextual knowledge, especially Languages (e.g., SEA languages and dialects such as \href{https://en.wikipedia.org/wiki/Malayo-Polynesian_languages}{Malayo-Polynesian}, \href{https://en.wikipedia.org/wiki/Wa_language}{Wa}, and \href{https://en.wikipedia.org/wiki/Cua_language_(Austroasiatic)}{Cua}) and Science and Technology (i.e., \href{https://en.wikipedia.org/wiki/Singtel}{Singtel}, \href{https://en.wikipedia.org/wiki/AXN_(Asian_TV_channel)}{AXN}, and \href{https://en.wikipedia.org/wiki/Kompas}{Kompas}), are consistently more challenging, with average F1 scores dropping to 58\%-62\%.
This suggests that SEA-NLI not only tests surface-level cultural recognition, but crucially also exposes weaknesses in models' ability to reason over culturally grounded and knowledge-intensive concepts.
Note that we demonstrate the category results using English texts in Appendix~\ref{apd:category_result_english}, where we also observe the same trend as SEA languages.

\section{Insights for Future Model Development} \label{sec:insights}

To answer \textbf{RQ3}, we conduct three studies to identify challenges and gaps, and to improve the model's ability to understand more SEA cultures.

\subsection{Understand the Challenge of Hard Set}

To explain the performance drop on the \emph{hard} set, we test whether success depends on explicit cultural keyword overlap.
Many \emph{hard} examples omit the target concept, so correct predictions require cultural understanding rather than surface lexical cues.
For instance, given a premise containing \textit{\href{https://en.wikipedia.org/wiki/Hokkien_mee}{Hokkien mee}} and a hypothesis mentioning ``noodles'', a model can predict entailment only if it knows that ``mee'' refers to ``noodles'' in this cultural context.
We therefore compare F1 between samples with and without concept occurrence in the premise or hypothesis.

As shown in Table~\ref{tab:keyword_ana}, models perform worse when the cultural concept appears in neither the premise nor the hypothesis.
When the concept is absent, models need to understand the concept behind those topics without relying on superficial cues.
In addition, we notice that the SEA-adapted model, Gemma-SEALION-27B, always performs better than the original model in all cases.
These results suggest that SEA adaptation injects useful regional knowledge into the model, making it more resilient to the lack of explicit cultural keywords.
Thus, the hard set is challenging because models must infer cultural background beyond surface-level lexical matching, but SEA-specialized models are better equipped to handle this problem.

\begin{table}[h!]
\centering
\vspace{-3mm}
\footnotesize 
\setlength{\tabcolsep}{3pt} 
\label{tab:gpt4_performance_simplified}
\begin{tabular}{llccc}
\toprule
\textbf{Model} & \textbf{Concept Occurrence} & \textbf{Count} & \textbf{SEA} & \textbf{ENG} \\ \midrule

 

\multirow{2}{*}{\rotatebox[origin=c]{90}{\textbf{\shortstack{\tiny{Gemma-}\\\tiny{3-27B-it}}}}} 
 & Both + Once & 256 & 77.41 & 77.66 \\
 & Neither & 461 & 76.51 & 75.90 \\
 \midrule\midrule

\multirow{2}{*}{\rotatebox[origin=c]{90}{\textbf{\shortstack{\tiny{Gemma-}\\\tiny{SEA-}\\\tiny{LION}\\\tiny{-v4-27B}}}}} 
 & Both + Once & 256 & 78.39 & 78.46 \\
 & Neither & 461 & 77.00 & 76.02	 \\
\bottomrule


\end{tabular}
\vspace{-3mm}
\caption{The analysis of hard set on cultural keyword overlapping using weighted F1 score.}
\label{tab:keyword_ana}
\vspace{-5mm}
\end{table}

\subsection{Culture-aware Prompting}
\label{subsec:cultural_aware_prompting}

LLMs are trained on large-scale corpora, including Wikipedia~\cite{grattafiori2024llama3herdmodels, gemmateam2025gemma3technicalreport,ng-etal-2025-sea}, which may contain substantial knowledge about SEA cultures.
However, such knowledge may remain latent and require targeted prompting to be reliably elicited.
We therefore study culture-aware prompting strategies for extracting SEA-specific knowledge from LLMs, including cultural persona (\texttt{Cult.}), topic summaries (\texttt{Sum.}), target topics (\texttt{Topic}), in-context examples (\texttt{ICL}), and reasoning tokens (\texttt{CoT}); full prompts are provided in Appendix~\ref{fig:base_prompt}.

In Table~\ref{tab:prompting}, \texttt{Cult.}+\texttt{Sum.} yields the largest gains, and other culture-enriched variants also improve SEA adaptation, suggesting that explicit SEA context helps models retrieve relevant knowledge.
In contrast, \texttt{CoT} does not improve performance in most cases.
Appendix~\ref{apd:Reduced performance on reasoning prompts} shows that reasoning prompts often shift models toward lexical overlap, leading to over-prediction of \texttt{neutral}.
%
For example, in Figure~\ref{fig:reasoning_prompt_error}, Gemma-SEA-LION-v4-27B understands the Thai uniform context but tries to map the premises keywords to the hypothesis.
Without a keyword match, it defaults to \texttt{neutral} instead of inferring the cultural concept.
We observe a similar pattern with \texttt{ICL}, where demonstrations encourage lexical matching over cultural understanding.
Overall, these results suggest that SEA-NLI calls for cultural-knowledge adaptation more than generic reasoning elicitation.

\begin{table}[h!]
\centering
\vspace{-3mm}
\setlength{\tabcolsep}{4pt} 
\scalebox{0.82}{
\begin{tabular}{cl cccc}
\toprule
& & \multicolumn{2}{c}{\textbf{\begin{tabular}[c]{@{}c@{}}Gemma-3\\ 27B-IT\end{tabular}}} & \multicolumn{2}{c}{\textbf{\begin{tabular}[c]{@{}c@{}}Gemma-SEA\\ LION-v4-27B\end{tabular}}} \\
\cmidrule(lr){3-4} \cmidrule(lr){5-6}
\textbf{Set} & \textbf{Method} & \textbf{SEA} & \textbf{ENG} & \textbf{SEA} & \textbf{ENG} \\
\midrule
\multirow{7}{*}{\rotatebox[origin=c]{90}{\textit{Normal Set}}} 
& Base             & 87.16 & 87.06 & 87.51 & 86.69 \\
& Base+Cult.          & 87.12 & 87.92 & 87.19 & 88.06 \\
& Base+Cult.+Topic  & 87.76 & 88.18 & 88.18 & 88.68 \\
& Base+Cult.+Sum.   & \textbf{89.99} & \textbf{91.46} & \textbf{90.48} & \textbf{91.46} \\
& CoT+Cult.              & 84.59 & 86.05 & 86.48 & 86.81 \\
& CoT+Cult.+Sum.           &  86.67 & 88.17  & 87.30 & 89.28 \\
& CoT+ICL+Cult.              & 85.31 & 85.96 & 85.94 & 86.70 \\
\midrule
\multirow{7}{*}{\rotatebox[origin=c]{90}{\textit{Hard Set}}} 
& Base             & 76.85 & 76.53 & 77.53 & 76.90 \\
& Base+Cult.          & 76.42 & 76.04 & 77.27 & 76.34 \\
& Base+Cult.+Topic  & 75.92 & 77.59 & 76.58 & 77.64 \\
& Base+Cult.+Sum.   & \textbf{77.59} & 78.15 & \textbf{77.85} & \textbf{78.44} \\
& CoT+Cult.             & 75.73 & 76.17 & 76.06 & 75.61 \\
& CoT+Cult.+Sum.           & 75.95 &  \textbf{78.17} & 77.68 & 78.28\\
& CoT+ICL+Cult.   & 76.28 & 75.56 & 76.57 & 75.48 \\
\bottomrule
\end{tabular}}
\vspace{-2mm}
\caption{Weighted F1-score performance (\%) for each prompt variant.}
\vspace{-5mm}
\label{tab:prompting}
\end{table}

\subsection{Error Analysis}
\begin{figure}[h!]
  \centering
  \vspace{-5mm}
   \includegraphics[width=1\columnwidth]{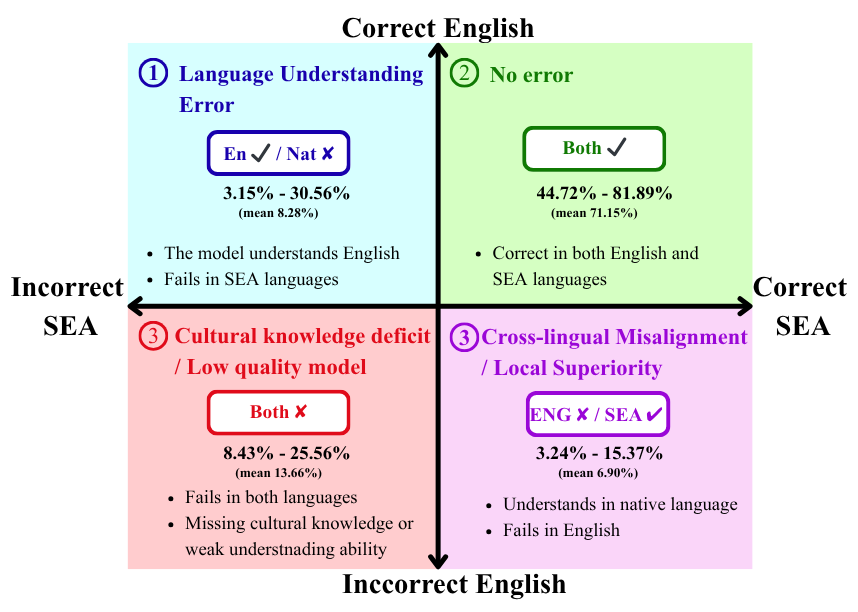}
  \vspace{-9mm}
  \caption{Error taxonomy mapping English and SEA prediction outcomes to distinct model deficiencies.}
  \vspace{-5mm}
  \label{fig:error-analysis-method}
\end{figure}

To analyze model failure cases, we compare predictions on parallel English and SEA language NLI samples using 17 models, as shown in Figure~\ref{fig:error-analysis-method}.
We group errors into three categories: \textit{language misunderstanding} (ENG $\checkmark$/SEA $\times$), \textit{cultural knowledge deficits} (ENG $\times$/SEA $\times$), and \textit{cross-lingual misalignment or translation loss} (ENG $\times$/SEA $\checkmark$).
The full results are shown in Table~\ref{tab:error_full}.
We summarize the results as follows.

\noindent
\textbf{Cultural Knowledge Deficits Dominate Model Errors.}
Most errors are cultural knowledge deficits, accounting for 8.43\%-25.56\% of cases across models.
Even top models show non-trivial rates ($\sim$9--11\%), suggesting that many failures stem from gaps in SEA-specific semantic and cultural knowledge rather than translation artifacts.
For example, on ``\href{https://en.wikipedia.org/wiki/Laksa}{Laksa},'' all models fail to distinguish sour and spicy laksa, which differ in taste and cooking method, and predict \texttt{neutral}.
Models that fail on English samples also typically fail on SEA-language samples, indicating shared deficiencies rather than language-specific noise.

\noindent
\textbf{SEA Language Understanding Remains a Major Bottleneck.}
Language misunderstanding is the second-largest error category, accounting for 3.15\%-30.56\% of errors.
This is especially common in smaller models such as LLaMA-3.1-8B (30.56\%), suggesting reliance on English-centric representations.
By contrast, cross-lingual misalignment remains low (3\%-8\%), indicating that models rarely succeed in SEA languages when failing in English.
Thus, model knowledge appears more accessible in English, limiting real-world SEA applications where users need reliable answers in their native languages.

\section{Conclusion}

We introduce SEA-NLI, a culturally grounded NLI benchmark for evaluating encoder- and decoder-based models in Southeast Asian contexts.
Our evaluation of 17 models shows that SEA-NLI remains challenging even for strong frontier and SEA-adapted models.
Our analysis suggests that this degradation is driven less by generic reasoning failures than by missing SEA cultural knowledge: models struggle when cultural concepts are implicit, while culture-aware prompting improves performance more reliably than chain-of-thought.
Cross-lingual error analysis further indicates that failures are dominated by cultural knowledge deficits and SEA language misunderstanding, rather than translation misalignment.
Together, these findings highlight the need for models adapted not only to Southeast Asian languages but also to Southeast Asian cultural knowledge.



\section*{Limitation}
This study is subject to limitations related to the full range of linguistic and cultural diversity across all Southeast Asia countries. 
The evaluation is restricted to a selected subset of languages and does not account for all dialects, registers, or code-switching practices. 
These constraints limit the generalization of the findings and suggest that further work is encouraged to develop more comprehensive and locally grounded NLI datasets.

Similar to other benchmark works, we did not present a new model that mitigates the SEA safety problem. 
However, we dedicate the whole Section~\ref{sec:insights} to how to achieve a high score on our benchmark.
We present both errors and cultural knowledge studies for future work that are interesting to work on the SEA NLI problems.
This is an insight for future works to improve the SEA cultural understanding. 

Moreover, since we use generative AI for data creation, there might be another bias in the dataset.
However, we did check with humans and previous works~\cite{gururangan-etal-2018-annotation,zhou-bansal-2020-towards,mccoy-etal-2019-right,belinkov-etal-2019-adversarial} and not found any patterns for models to find a shortcut in our dataset, i.e., negation rule, length prediction, lexicon matching, semantic matching, etc., as shown in Appendix~\ref{appendix:bias}.

\section*{Ethical Statement}

All annotators were native speakers of SEA languages and were of legal working age (above 18 years old). 
The majority of annotators were students enrolled at local universities and participated on a voluntary, compensated basis. 
They were recruited through a rigorous screening and evaluation process. Prior to their participation, they received comprehensive training to ensure informed, accurate, and responsible contributions. 
Annotators were compensated at a rate of \$17 USD per hour, which exceeds the typical market rate for comparable annotation work, reflecting a commitment to fair and ethical labor practices.
In addition, we also have an IRB for this project, allowing us to leverage the annotator to help us check and revise the cultural datasets for SEA languages. 

\section*{Acknowledgments}
This project is supported by the National Research Foundation, Singapore under its National Large Language Models Funding Initiative. Any opinions, findings and conclusions or recommendations expressed in this material are those of the author(s) and do not reflect the views of National Research Foundation, Singapore.

\bibliography{custom,filtered_merged}


\clearpage

\appendix

\section*{Appendix}
\label{apd:dataset_statistic}

\section{Data Statistics} \label{subsec:data_statistic}
\begin{figure}[h!]
  \centering
  \vspace{-3mm}
  \includegraphics[width=\linewidth]{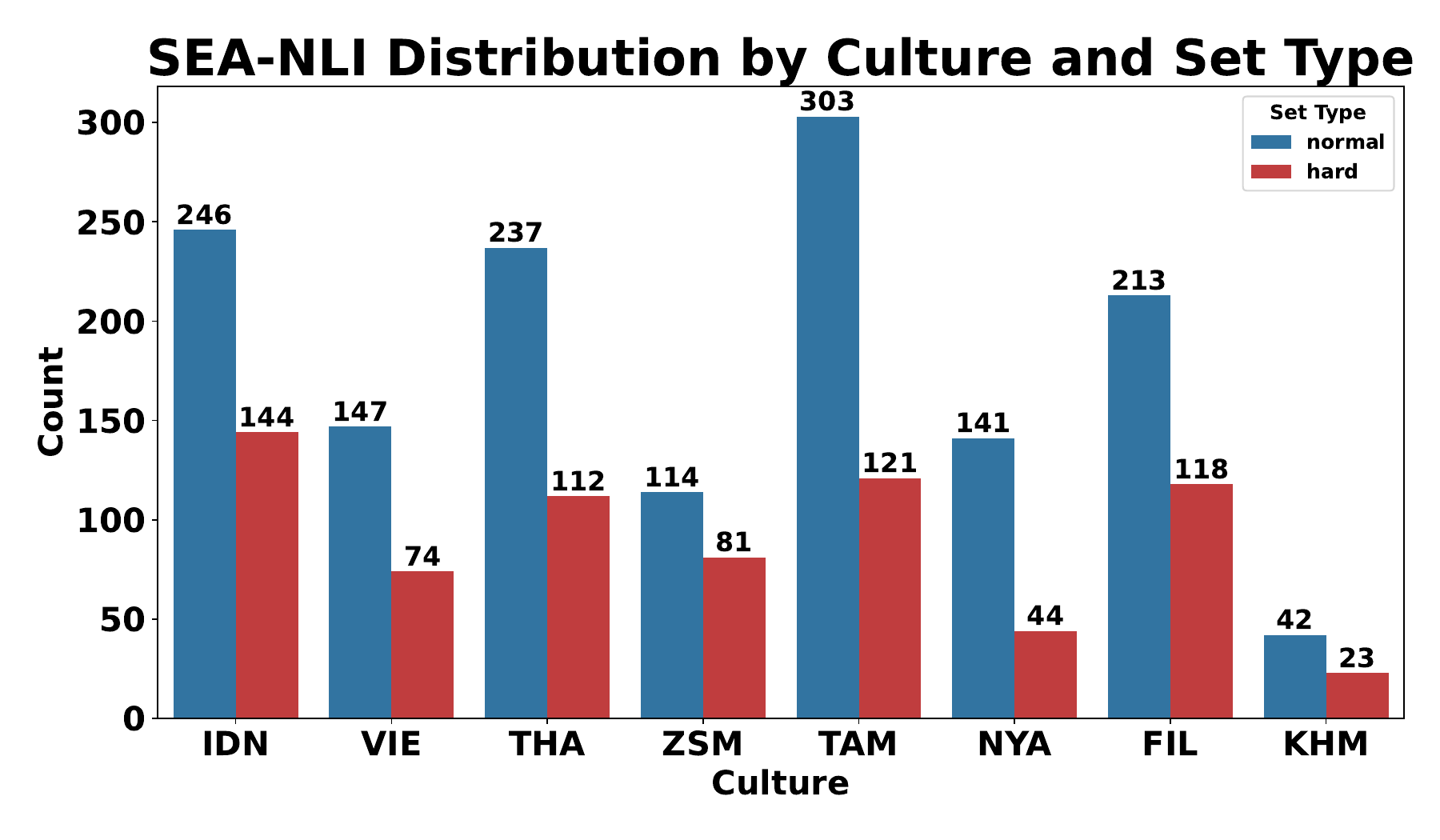}
  \vspace{-8mm}
  \caption{ Distribution of SEA-NLI samples across eight cultures, partitioned into normal and hard subsets.}
  \vspace{-3mm}
  \label{fig:SEA-NLI-distribution-culture}
\end{figure}

\begin{figure}[h!]
  \centering
  \vspace{-3mm}
  \includegraphics[width=\linewidth]
  {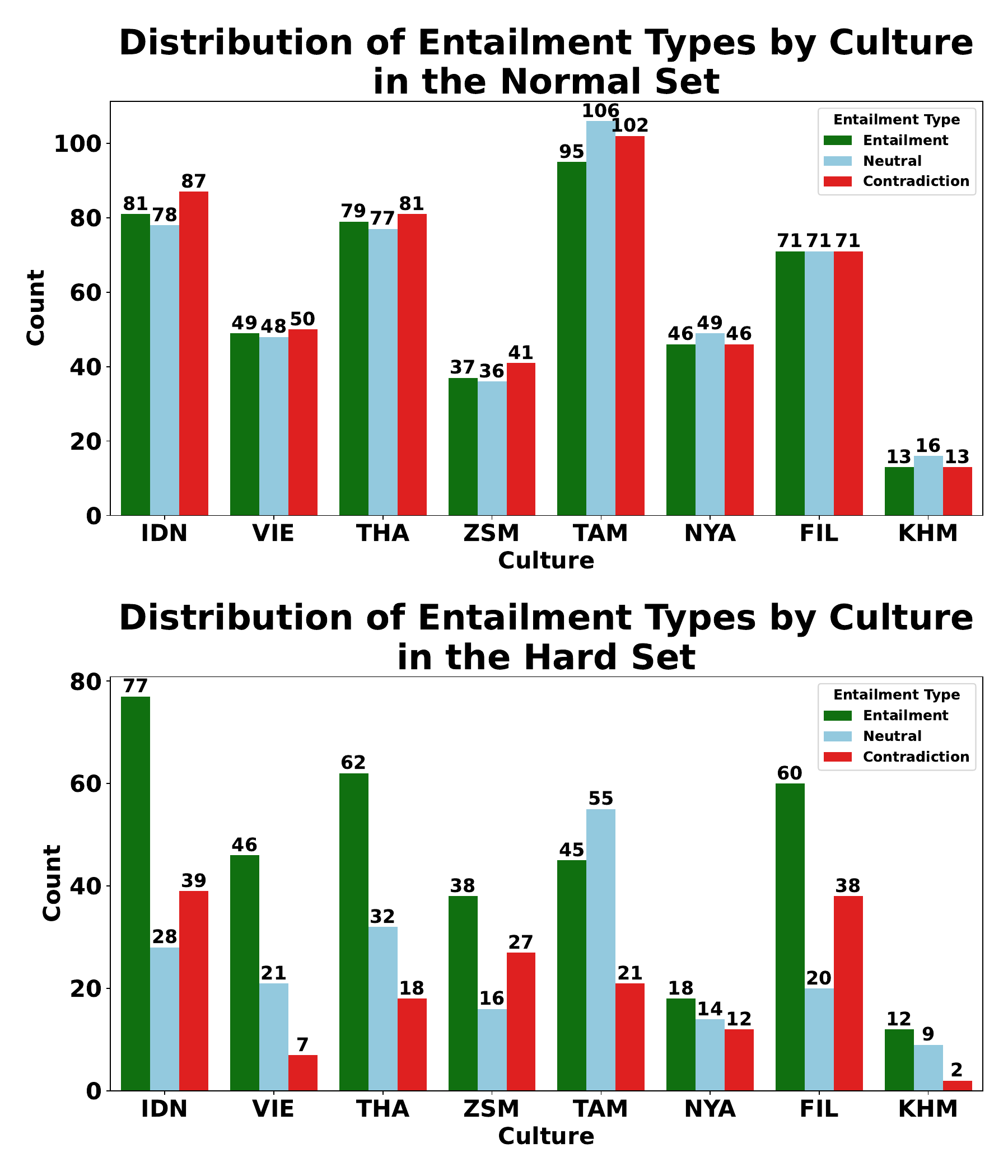}
  \vspace{-8mm}
  \caption{Comparison of Entailment Type Distributions between Normal and Hard Sets in the SEA-NLI Dataset by Country.}
  \vspace{-3mm}
  \label{fig:SEA-NLI-distribution-entaiment-type}
\end{figure}

This section summarizes the key statistics of the SEA-NLI dataset. 
The dataset exhibits a long-tailed distribution; while Singapore (TAM) and Indonesia (IDN) provide the highest sample volumes, the inclusion of smaller subsets such as Cambodia ensures broad regional representation. Across all cultures, the normal set consistently maintains a higher volume than the hard set, typically following a 2:1 ratio as shown in Figure~\ref{fig:SEA-NLI-distribution-culture}.
In addition, Figure~\ref{fig:SEA-NLI-distribution-entaiment-type} shows that the normal subset has balanced entailment labels, while the hard subset exhibits greater variance across cultures.

\section{Preliminary Pilot Set}
\label{apd:pilot_generation_prompts}
\begin{figure}[h!]
  \centering
  \includegraphics[width=1\columnwidth]{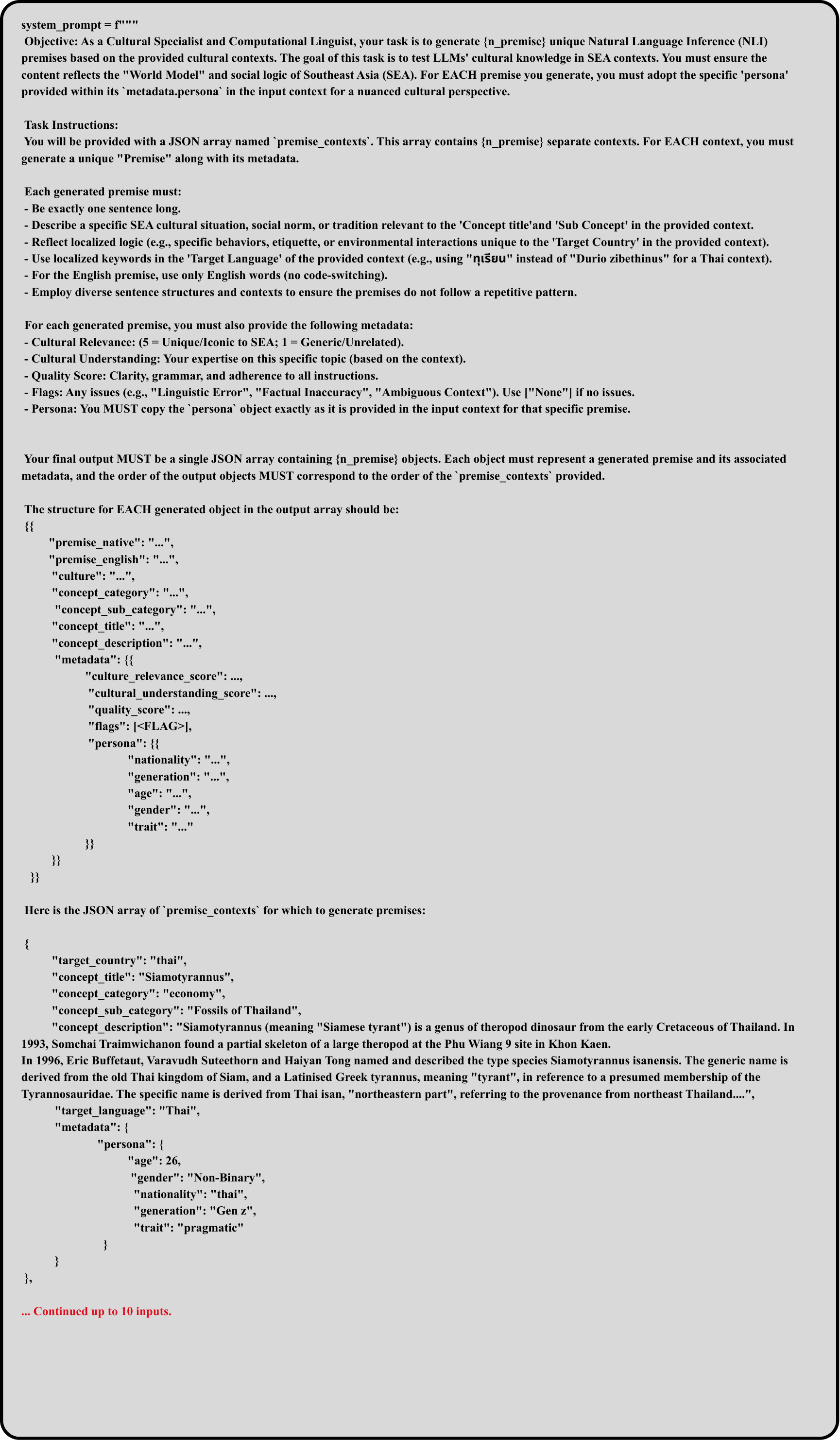} 
  \vspace{-3mm}
  \caption{The premise generation prompt for the pilot dataset.}
  \label{fig:premise-generation-prompt_batch1}
  \vspace{-0.5em}
\end{figure}

\begin{figure}[h!]
  \centering
  \includegraphics[width=1\columnwidth]{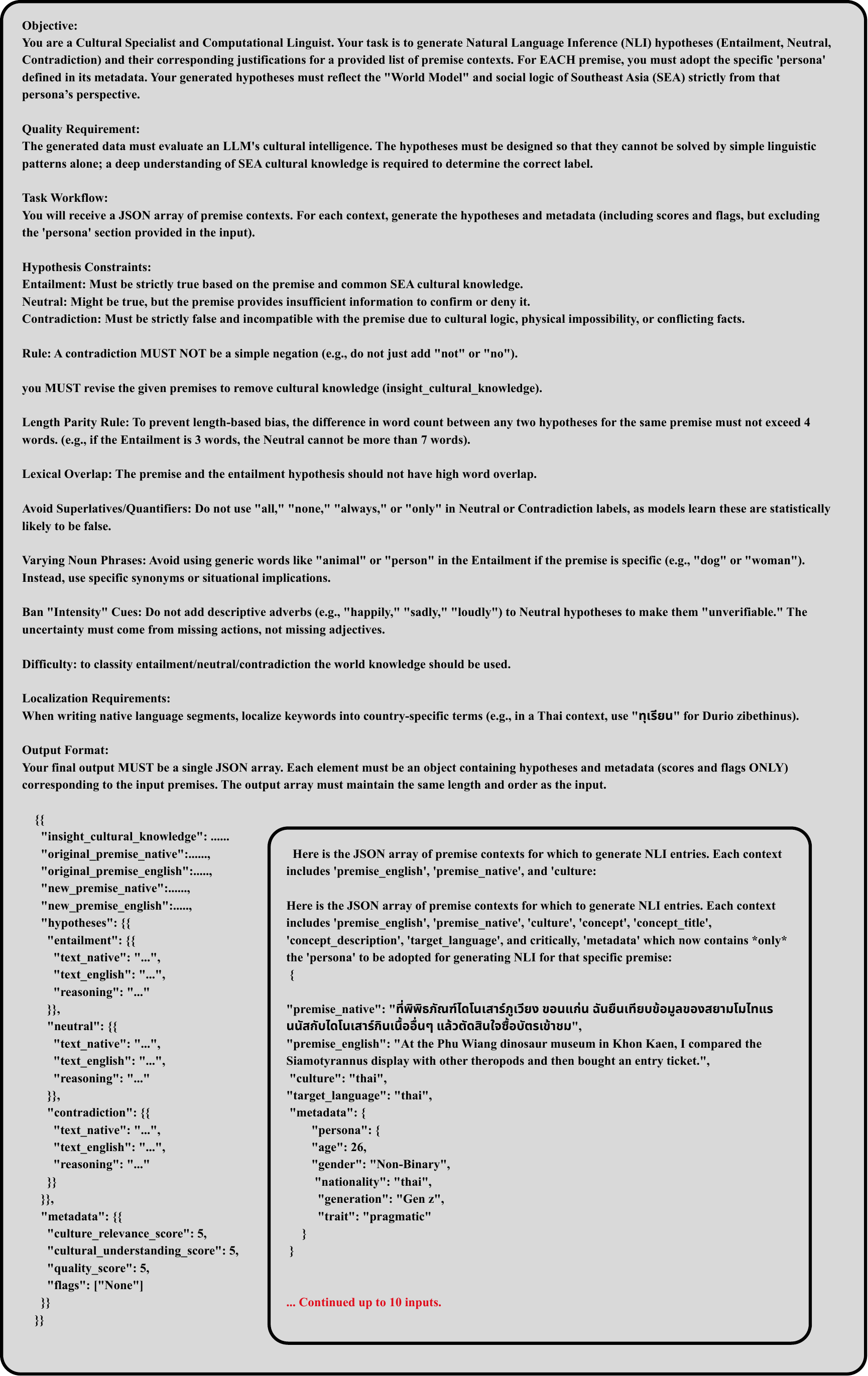} 
  \vspace{-3mm}
  \caption{The hypothesis generation prompt for the pilot dataset.}
  \label{fig:hypothesis-generation-prompt_batch1}
\end{figure}

\begin{figure*}[h!]
  \centering
  \includegraphics[width=2\columnwidth]{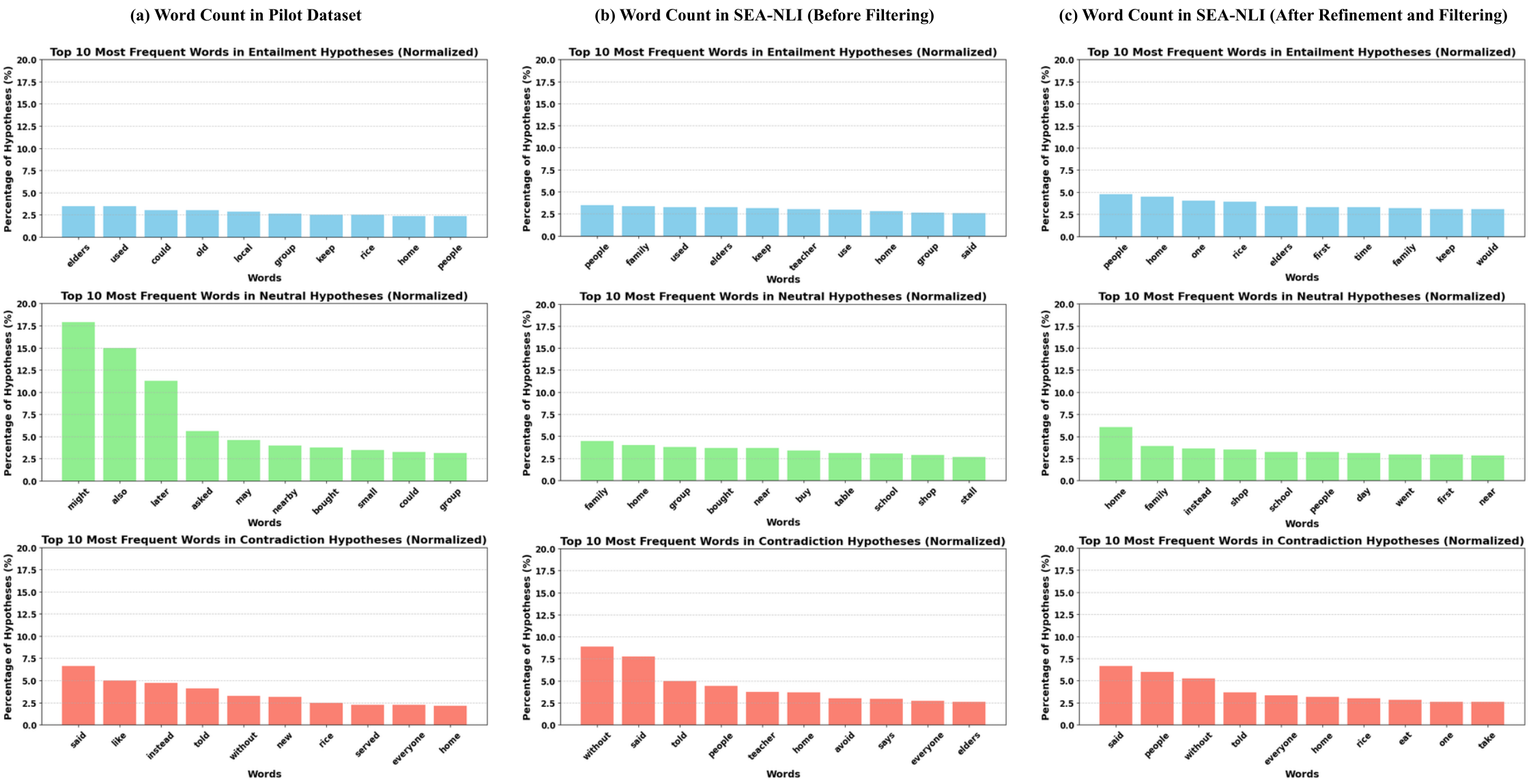} 
  \vspace{-3mm}
  \caption{Evolution of the SEA-NLI word frequency distribution. All values are normalized to the percentage of total hypotheses per label. The sharp spikes in the pilot dataset (a) were addressed through prompt engineering (b) and automated refinement and filtering (c), resulting in a balanced and robust distribution of concrete semantic terms across all entailment categories.}
  \vspace{-3mm}
  \label{fig:word_count_all}
\end{figure*}

To assess the quality of the LLM-generated NLI samples (GPT-5.2), we conduct a study using the premise and hypothesis generation prompts in Figures~\ref{fig:premise-generation-prompt_batch1} and~\ref{fig:hypothesis-generation-prompt_batch1}, respectively.
Initial analysis of 2,400 premise--hypothesis pairs revealed that frequent words were spuriously associated with specific NLI labels, introducing potential annotation artifacts (Figure~\ref{fig:word_count_all}a).
To mitigate these biases, we refined the prompting strategy, as detailed in section~\ref{apd:dataset_generation_prompts}.
Moreover, we sampled 120 premise--hypothesis pairs per culture for human evaluation.
Detailed pilot results regarding inter-annotator agreement, cultural relevance, and linguistic naturalness are provided in Appendices~\ref{apd:annotation_agreement}.

\section{Iterative Refinement and Dataset Improvement}
\label{apd:dataset_improvement_prompts}
To enhance the quality of the SEA-NLI dataset, we employ an iterative refinement loop using GPT-5.2, as detailed in Section~\ref{subsec:data_filtering}. 
This process is designed to mitigate annotation artifact bias.
\subsection{Regeneration Prompt}
\label{subsec:regeneration_prompt}

\begin{figure}[h!]
  \centering
  \includegraphics[width=\columnwidth]{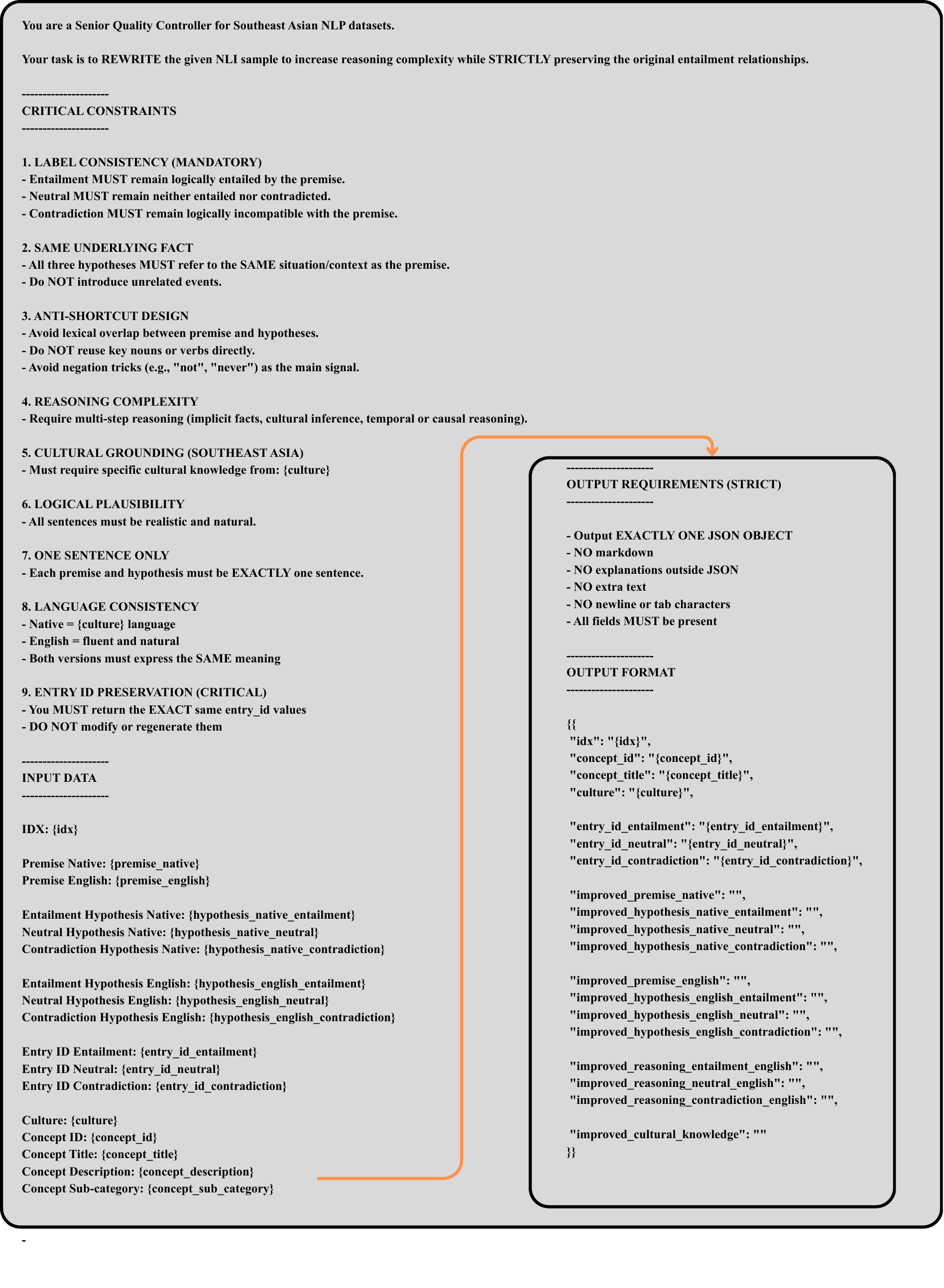} 
  \vspace{-5mm}
  \caption{The regeneration prompt for improving NLI sample quality in each refinement loop.}
  \vspace{-3mm}
  \label{fig:regeneration_prompt}
\end{figure}

Figure~\ref{fig:regeneration_prompt} illustrates the prompt template used for sample regeneration.
The prompt accepts a premise, its corresponding hypotheses in both English and Southeast Asian languages, and specific concept metadata.
The prompt also enforces rigorous quality control by anchoring all hypotheses to a single underlying fact and ensuring logical consistency.
To maximize robustness, an ``anti-shortcut'' design is utilized; this strategy penalizes lexical overlap, necessitates deep regional cultural grounding, and mandates that all sentences remain logically plausible and concise.
%

\subsection{What Did We Get After the Refinement Process?}

\begin{figure}[h!]
  \centering
  \vspace{-3mm}
  \includegraphics[width=\linewidth]{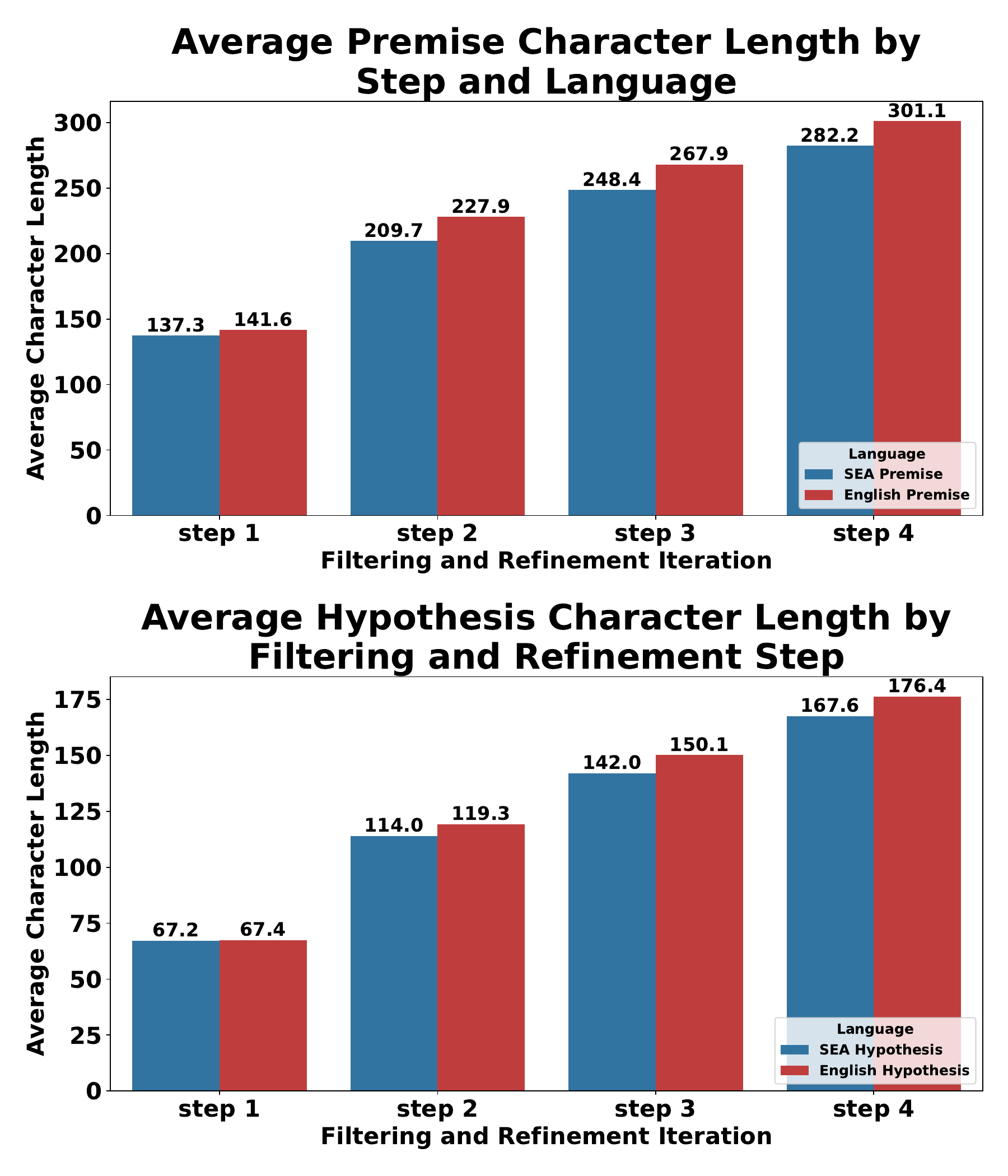}
  \vspace{-8mm}
  \caption{Average character length of premises (top) and hypotheses (bottom) across LLM filtering and refinement step (Steps 1-4) for both SEA and English language samples.}
  \vspace{-3mm}
  \label{fig:SEA-NLI-distribution-length-in-each-step}
\end{figure}

The refinement process results in a measurable growth in sample length. As shown in Figure~\ref{fig:SEA-NLI-distribution-length-in-each-step}, the average character length of premises and hypotheses accepted at each stage increases consistently from Step 1 through Step 4 across all languages. 
Moreover, the length that increased also means the explanation that increased for each premise--hypothesis pair. 
As shown in Figure~\ref{fig:compare-dataset} in the main text, the hard sample has a longer context than the normal sample since it tries to explain the concept of each cultural topic.
%

\begin{figure}[h!]
  \centering
  \includegraphics[width=1\linewidth]
  {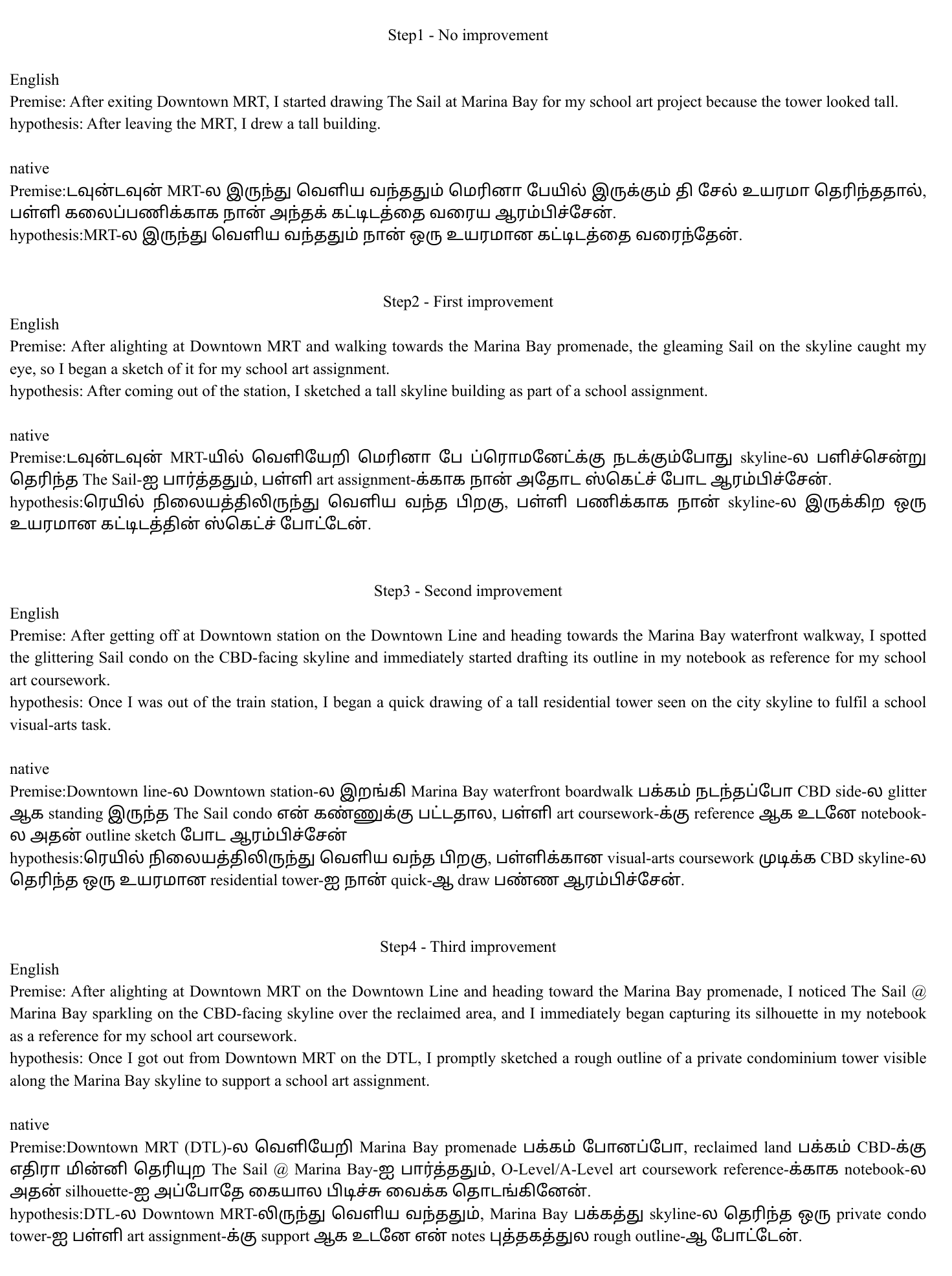}
  \vspace{-8mm}
  \caption{Example of SEA-NLI in each regeneration step.}
  \vspace{-5mm}
  \label{fig:nli-improvement-in-each-step.pdf}
\end{figure}

\begin{figure}[h!]
  \centering
  \includegraphics[width=1\linewidth]
  {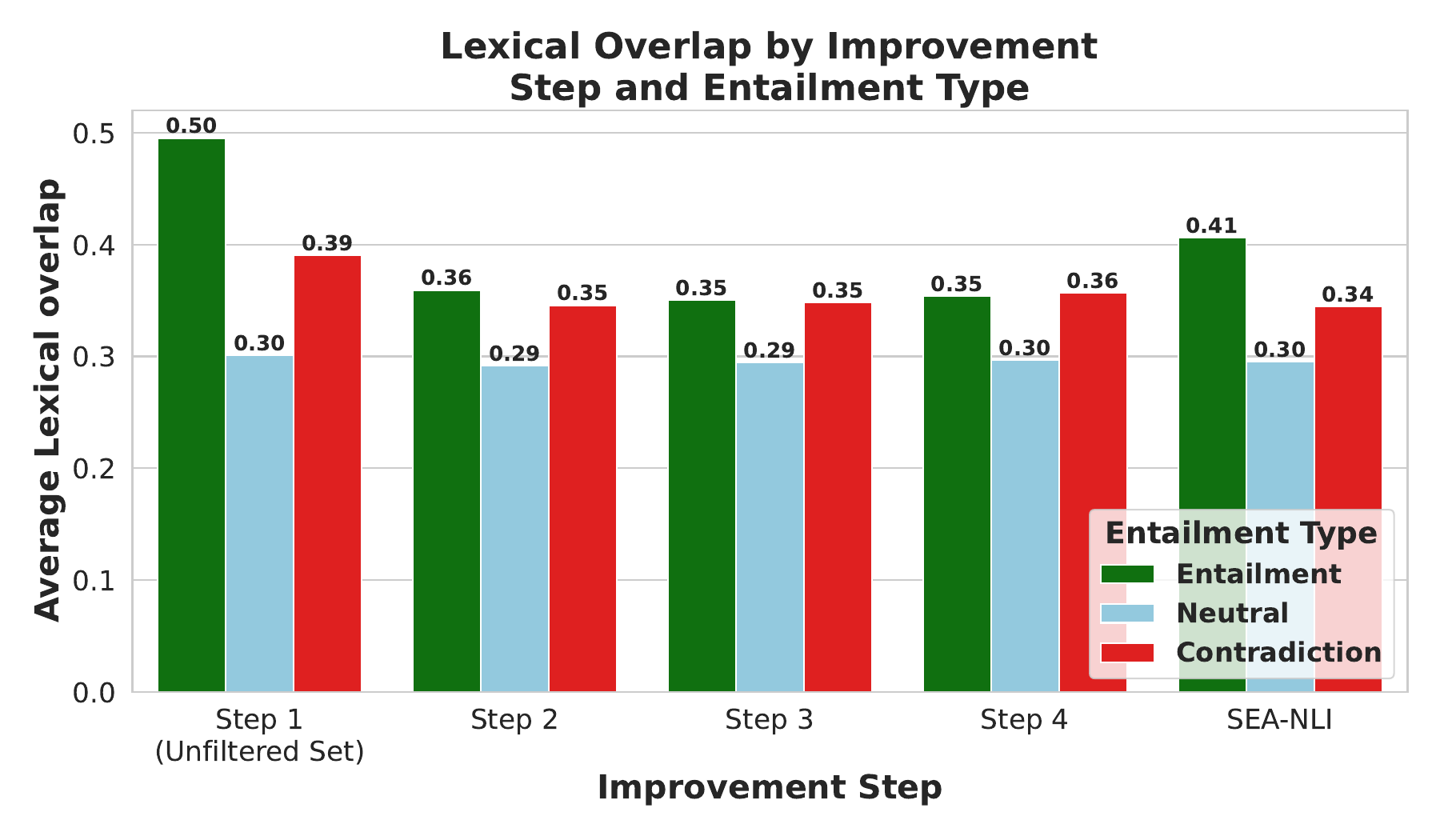}
  \vspace{-8mm}
  \caption{Lexical overlap between premise and hypothesis in each regeneration step.}
  \vspace{-5mm}
  \label{fig:overlap_improvement_step}
\end{figure}

Beyond length, the iterative loop actively removes annotation bias. Figure~\ref{fig:nli-improvement-in-each-step.pdf} provides a qualitative example of how a sample evolves, becoming harder to classify without deep semantic understanding. Quantitatively, this is supported by a reduction in lexical shortcuts: the loop balances word overlap across entailment types (Figure~\ref{fig:overlap_improvement_step}) and eliminates class-specific "trigger words" that models often exploit as heuristics (see Figure~\ref{fig:word_count_all}c).


\section{Inference Speed vs. Performance Analysis}
\begin{figure*}[h!]
  \centering
  \vspace{-3mm}
  \includegraphics[width=\linewidth]{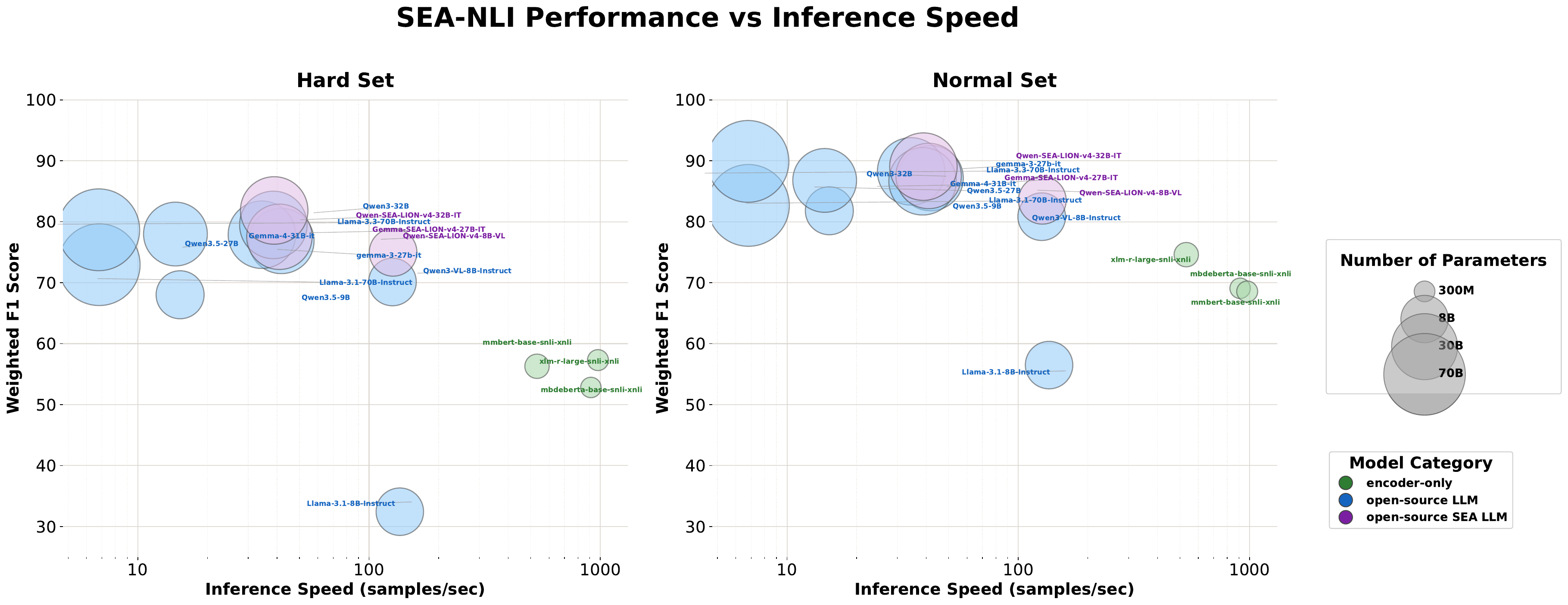}
  \vspace{-8mm}
  \caption{Inference speed vs. F1 score on the SEA-NLI benchmark. Results are shown for SEA performance (left) and English performance (right). Bubble sizes scale with parameter count, illustrating the trade-off between model scale, speed, and F1-score.}
  \vspace{-3mm}
  \label{fig:test-time-inference}
\end{figure*}

The relationship between model performance on the SEA-NLI dataset and inference throughput is illustrated in Figure~\ref{fig:test-time-inference}. 
The graph illustrates the efficiency-performance trade-off across different architectures on the SEA-NLI benchmark. 
While encoder-only models (green) offer superior throughput, they lack the cultural understanding found in larger LLMs. 
A key finding is the ``cultural collapse'' of general models like Llama-3.1-8B-Instruct, which performs well in English but fails in the SEA context. 
Conversely, SEA-specific LLMs (purple) demonstrate high cultural robustness, maintaining top-tier F1 scores in regional contexts despite lower inference speeds. 
This underscores that for regional tasks, localized training is more critical for robustness than sheer inference speed or general-purpose scaling.

\section{Annotator Agreement and Details}
\label{apd:annotator_details}

To ensure linguistic and cultural authenticity, we established a rigorous quality control framework involving human experts. 
This section outlines the selection criteria for Southeast Asian language specialists, the metrics employed to evaluate LLM outputs, and the methodologies for measuring inter-annotator agreement. 
By combining expert oversight with standardized scoring for cultural nuance and grammatical precision, we measured inter-annotator agreement specifically for the NLI labels to confirm logical consistency. Furthermore, we assessed cultural relevance, cultural understanding, and grammatical precision through expert-assigned quality scores. 

\subsection{Annotator Selection}

To select annotators for the study, we recruit native speakers and language experts from eight Southeast Asian countries, where we give two tasks to annotators. 
The first task is entailment type classification, which includes twelve multiple-choice questions where annotators identify if a relationship is an Entailment, Neutral, or Contradiction (12 examples). 
The second task is hypothesis generation, which includes three questions where annotators are given a premise and must write their own entailment, neutral, and contradiction hypotheses (3 examples). 
%
%
We select only the annotator that pass the tasks with a score of more than 80\%.
This process confirms that every annotator fully understands the logic required for the project. 

\subsection{Annotator for Cultural Filtering} \label{appendix:screening}
For the cultural filtering process in Section~\ref{subsec:concept_garthering}, we asked three annotators for each country to filter unrelated cultures.
In particular, we asked annotators to check whether they were familiar with the concept title from scraped Wikipedia or not, where we removed only the topics that all three annotators were unfamiliar with.
We also found that the rejection rate is only $\sim$5\% from around $\sim$24,000 culture topics.
The pattern we found is that the culture topics that do not have much information on Wikipedia pages would be removed. 

Then, we filter the topic that is the least popular in Wikipedia using n\_link, where the number indicates the number of citations in Wikipedia.
We removed the page that has the least 70\% citation in Wikipedia, where the number was also calculated by our annotators, increasing or decreasing the number might include the least popular or not well-known topics.
Using these topics might be incorrect in the label verification step since most annotators might not fully understand them.

\subsection{Human Evaluation Metrics}
\label{apd:human_evaluation_metrics}
To evaluate the quality of LLM-generated NLI samples in Section~\ref{subsec:qa}, we utilize the following metrics:

\begin{enumerate}
    \item \textbf{Cultural Relevance Score:} Adapted from \citet{cahyawijaya-etal-2025-crowdsource}, this metric assesses how effectively the generated content aligns with the intended Southeast Asian (SEA) cultural context.
    \begin{compactitem}[\hspace{\setalign}•]
        \item \textbf{Score 5 (Unique to SEA):} The premise describes traditions, objects, or landmarks that originate in SEA and are considered iconic, such as Pad Thai, Batik, Songkran, or the Petronas Towers.
        \item \textbf{Score 3 (Ubiquitous):} The content features concepts that are not exclusive to SEA but are common in daily life or possess regional nuances, such as the Chinese New Year in Singapore, local mosques, or specific tropical fruit flavors.
        \item \textbf{Score 1 (Unrelated):} The content involves concepts unrelated to SEA, such as the Super Bowl, the Statue of Liberty, or generic international brands like IKEA.
    \end{compactitem}

     Intermediate scores (2 or 4) should be assigned if the context falls between these primary categories.
     
    \item \textbf{Cultural Understanding Score:} This score quantifies the annotator's personal familiarity with the specific cultural context of the sample.
    \begin{compactitem}[\hspace{\setalign}•]
        \item \textbf{Score 5 (Native / Expert):} The annotator identifies with the culture and understands specific nuances, local slang, and social traditions instinctively.
        \item \textbf{Score 4 (High Familiarity):} The annotator has lived in the region or possesses strong personal experience and comfort with the context.
        \item \textbf{Score 3 (Moderate Awareness):} The annotator recognizes the topic and its general meaning but lacks deep personal experience or specific detail.
        \item \textbf{Score 2 (Limited Exposure):} The annotator has heard of the tradition but has minimal knowledge of its actual practices.
        \item \textbf{Score 1 (No Prior Knowledge):} The culture is foreign to the annotator, who must rely entirely on the provided metadata to complete the task.
    \end{compactitem}

    \item \textbf{Quality Score:} This metric evaluates the linguistic clarity and contextual accuracy of the premise and its associated metadata.
    \begin{compactitem}[\hspace{\setalign}•]
        \item \textbf{Score 5 (Excellent):} The sentence is natural, grammatically perfect, and provides a clear cultural context.
        \item \textbf{Score 4 (Good):} The sentence is clear and usable, with only minor stylistic awkwardness.
        \item \textbf{Score 3 (Fair):} The sentence is understandable, but the phrasing is unnatural or the context is slightly vague.
        \item \textbf{Score 2 (Poor):} Significant grammatical issues or confusing context make logical deduction difficult.
        \item \textbf{Score 1 (Broken):} The sample is nonsensical, contains major factual errors, or is unusable.
    \end{compactitem}

    \item \textbf{Flagging Issues:} Annotators identify specific qualitative concerns that may impact the reliability of the sample. These include:
    \begin{compactitem}
        \item \textbf{Linguistic Error:} Significant grammar, spelling, or translation issues.
        \item \textbf{Factual Inaccuracy:} The premise contains incorrect information regarding the culture or location.
        \item \textbf{Ambiguous Context:} The statement is too vague to determine a definitive logical relationship.
        \item \textbf{Culturally Offensive:} The content is disrespectful or promotes harmful stereotypes.
        \item \textbf{Poor Formatting:} Issues with character encoding (e.g., broken local scripts) or punctuation.
        \item \textbf{Irrelevant to Topic:} The sample doesn't actually relate to the specific cultural topic or metadata provided.
        \item \textbf{Translation Mismatch:} The English translation and the native language sentence do not convey the same meaning.

    \end{compactitem}
\end{enumerate}

\subsection{Annotator Metrics}
\label{apd:annotation_agreement}

\subsubsection{Annotator Agreement}
Inter-annotator agreement was assessed using the Intraclass Correlation Coefficient (ICC) to validate the reliability and consistency of the human judgments \cite{shrout1979intraclass}. 
As shown in Table~\ref{tab:icc_results_updated}, we calculated the ICC for each cultural category individually to account for regional variances in perception. To derive a unified reliability metric across the dataset, these individual coefficients were aggregated using Fisher's Z-transformation \cite{fisher1921probable}. This transformation normalizes the distribution of the correlation coefficients, allowing for a statistically valid averaging process that accurately reflects the overall stability of the annotations across the diverse Southeast Asian languages studied. 

\begin{table}[ht]
\centering
\scalebox{0.7}{
\begin{tabular}{lccc}
\toprule
\textbf{Culture} & \textbf{Pilot Set} & \textbf{Non-filter Set} & \textbf{SEA-NLI} \\ \midrule
Cambodian        & 0.78             & 0.99             & 0.93             \\
Filipino         & 0.99             & 0.98             & N/A              \\
Indonesian       & 0.98             & 0.99             & 0.97             \\
Malaysian        & 0.96             & 1.00             & 0.90             \\
Myanmar          & 0.75             & 1.00             & 0.82             \\
Singaporean      & N/A              & N/A              & 0.93             \\
Thai             & 0.96             & 1.00             & 0.99             \\
Vietnamese       & 0.94             & 0.99             & 0.98             \\ \midrule
\textbf{Fisher’s Z ICC} & \textbf{0.95}    & \textbf{0.99}    & \textbf{0.96}    \\ \bottomrule
\end{tabular}}
\vspace{-2mm}
\caption{Inter-annotator agreement (ICC(C,k)) for Entailment Type Classification across three refinement rounds.}
\vspace{-3mm}
\label{tab:icc_results_updated}
\end{table}

\subsubsection{Cultural Relevant}
\label{apd:cultural_relevant}
Annotators assessed regional authenticity using the Cultural Relevance Score (CRS). As shown in Table~\ref{tab:cultural_relevance}, SEA-NLI achieved a high weighted average of 4.493, indicating that the samples are strongly grounded in Southeast Asian regional contexts.

\begin{table}[ht]
\centering
\small
\setlength{\tabcolsep}{4pt}
\scalebox{0.9}{
\begin{tabular}{lcccc}
\toprule
 & \textbf{Pilot Set} & \textbf{Unfiltered Set} & \multicolumn{2}{c}{\textbf{SEA-NLI}} \\ \cmidrule(lr){2-5}
\textbf{Culture} & ($n=120$) & ($n=300$) & \textbf{$n$} & \textbf{CRS} \\ \midrule
Cambodian   & 4.100 & 4.445 & 65  & 4.277 \\
Filipino    & 4.267 & 4.463 & 331 & 4.979 \\
Indonesian  & 4.505 & 4.458 & 390 & 4.599 \\
Malaysian   & 3.822 & 4.608 & 195 & 4.192 \\
Myanmar     & 4.558 & 4.825 & 185 & 4.881 \\
Singaporean & 3.875 & 3.017 & 424 & 3.692 \\
Thai        & 4.475 & 4.596 & 349 & 4.704 \\
Vietnamese  & 3.602 & 4.498 & 221 & 4.607 \\ \midrule
\textbf{Weighted Avg.} & \textbf{4.151} & \textbf{4.364} & \textbf{2160} & \textbf{4.493} \\ \bottomrule
\end{tabular}}
\vspace{-2mm}
\caption{Cultural Relevant Score Progression Across Evaluation Rounds}
\label{tab:cultural_relevance}
\vspace{-3mm}
\end{table}

\subsubsection{Naturalness}
\label{apd:naturalness_result}
The linguistic naturalness of the SEA-NLI dataset was evaluated by human annotators using the 'Quality Score' metric (Section~\ref{apd:human_evaluation_metrics}), which measures grammatical correctness, cultural authenticity, and stylistic fluency. 
As shown in Table~\ref{tab:quality_score}, from these sets, the overall weighted average improved from 4.151 to 4.651. While scores increased across the board, the relative rankings of individual cultures shifted significantly in each round, reflecting varying rates of progression. 
These results indicate a high level of quality and strong linguistic reliability, with final scores across all regional subsets approaching a perfect rating.
\begin{table}[ht]
\centering
\small
\setlength{\tabcolsep}{4pt}
\scalebox{0.9}{
\begin{tabular}{lcccc}
\toprule
 & \textbf{Pilot Set} & \textbf{Unfiltered Set} & \multicolumn{2}{c}{\textbf{SEA-NLI}} \\ \cmidrule(lr){2-5}
\textbf{Culture} & ($n=120$) & ($n=300$) & \textbf{$n$} & \textbf{Quality} \\ \midrule
Cambodian   & 4.079 & 4.477 & 65  & 3.623 \\
Filipino    & 4.719 & 4.982 & 331 & 4.983 \\
Indonesian  & 4.757 & 4.796 & 390 & 4.824 \\
Malaysian   & 4.200 & 4.661 & 195 & 4.403 \\
Myanmar     & 4.119 & 4.680 & 185 & 4.832 \\
Singaporean & 4.142 & 4.117 & 424 & 4.272 \\
Thai        & 4.467 & 4.582 & 349 & 4.772 \\
Vietnamese  & 4.373 & 4.651 & 221 & 4.756 \\ \midrule
\textbf{Weighted Avg.} & \textbf{4.357} & \textbf{4.618} & \textbf{2160} & \textbf{4.651} \\ \bottomrule
\end{tabular}}
\vspace{-2mm}
\caption{Quality Score Progression Across Evaluation Rounds}
\vspace{-3mm}
\label{tab:quality_score}
\end{table}


\subsubsection{Flags During Quality Control}
\begin{figure}[h!]
  \centering
  \vspace{-3mm}
  \includegraphics[width=\linewidth]{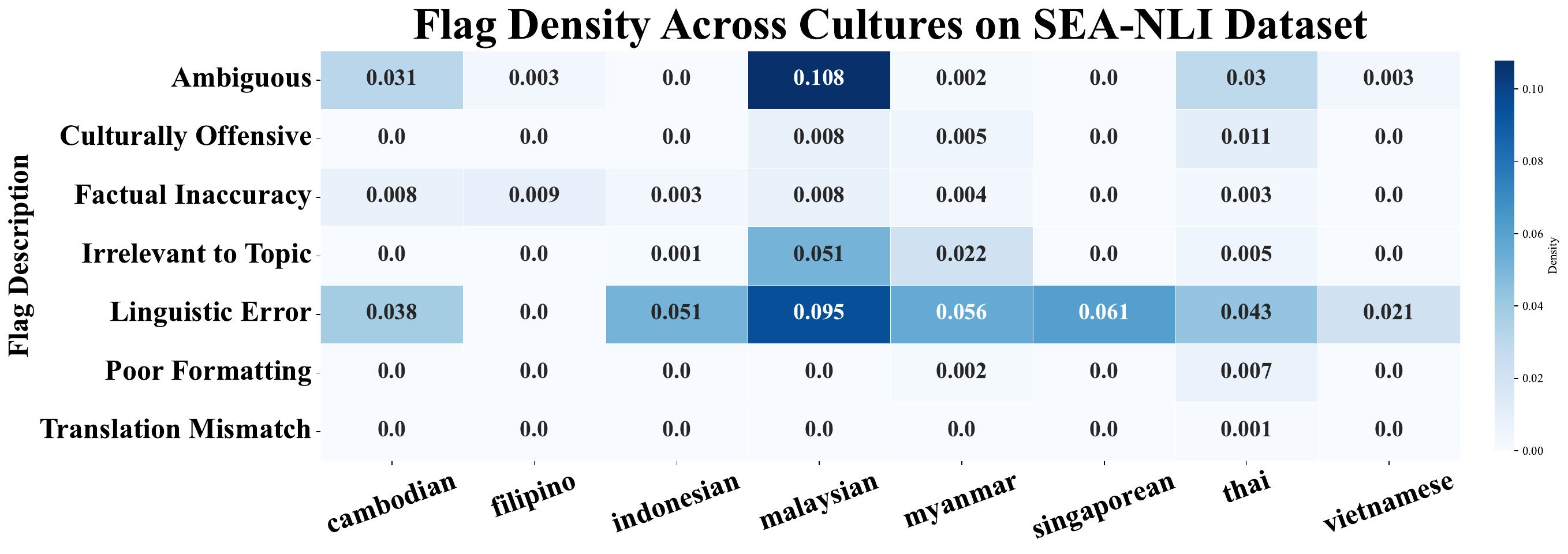}
  \vspace{-8mm}
  \caption{Distribution of data quality flags by culture on the SEA-NLI dataset. Higher values indicate a greater prevalence of specific issues, such as linguistic errors or ambiguity, within a given cultural subset.}
  \vspace{-3mm}
  \label{fig:flag-distribution}
\end{figure}

Samples flagged by annotators were adjudicated via majority vote to confirm the presence of anomalies. The proportions of anomalous samples are 8.75\%, 4.12\%, and 1.48\%, respectively. The distribution of raw flags prior to the majority voting process of SEA-NLI is presented in Figure~\ref{fig:flag-distribution}. The most frequent flag in this dataset is the linguistic error, which exhibits the highest overall density across the majority of cultures, followed by ambiguity.

\subsection{Annotator Details}
The annotators are native speakers of the target Southeast Asian languages and originate from eight Southeast Asian countries. The number of annotators assigned to each cultural group is detailed in Table \ref{tab:annotator_distribution}.


\begin{table}[ht]
\centering
\resizebox{\columnwidth}{!}{
\begin{tabular}{@{}llccc@{}} 
\hline
\textbf{Language} & \textbf{Country} & \multicolumn{3}{c}{\textbf{Count}} \\ \cline{3-5} 
                  & \textbf{of Origin} & \textbf{Pilot Set} & \textbf{Unfiltered Set} & \textbf{SEA-NLI} \\ \hline
Indonesian        & Indonesia                  & 5              & 6                  & 5                \\
Khmer             & Cambodia                   & 2              & 2                  & 2                \\
Malay             & Malaysia                   & 3              & 2                  & 2                \\
Myanmar           & Myanmar                    & 3              & 2                  & 3                \\
Filipino          & Philippines                & 3              & 2                  & 1                \\
Tamil             & Singapore                  & 1              & 1                  & 2                \\
Thai              & Thailand                   & 2              & 8                  & 8                \\
Vietnamese        & Vietnam                    & 4              & 4                  & 4                \\ \hline
\textbf{Total}    &                            & \textbf{23}    & \textbf{27}        & \textbf{27}      \\ \hline
\end{tabular}%
}
\vspace{-2mm}
\caption{Distribution of native-speaking annotators by language and country of origin.}
\vspace{-3mm}
\label{tab:annotator_distribution}
\end{table}

\section{Error Analysis Full Results}

The full error analysis results from all models are shown in Table~\ref{tab:error_full}. 

\begin{table}[h!]
    \centering
    \footnotesize
    \setlength{\tabcolsep}{2pt}
    \scalebox{0.7}{
    \begin{tabular}{p{4.5cm} cccc}
        \toprule
        & \textbf{Both} & \textbf{En $\checkmark$ /} & \textbf{En $\times$ /} & \textbf{Both} \\
        \textbf{Model} & \boldmath$\checkmark$ & \textbf{SEA $\times$} & \textbf{SEA $\checkmark$} & \boldmath$\times$ \\
        \midrule
        \textit{Encoder Models} & & & & \\
        mmBERT-base & 51.76 & 11.99 & 12.59 & 23.66 \\
        mDeBERTa-v3-base & 48.89 & 14.17 & 11.39 & 25.56 \\
        XLM-RoBERTa-large & 54.17 & 7.18 & 13.56 & 25.09 \\
        \midrule
        \textit{Decoder Models} & & & & \\
        Llama-3.1-8B-Instruct & 44.72 & 30.56 & 6.16 & 18.56 \\
        Qwen3-VL-8B-Instruct & 68.56 & 9.31 & 8.24 & 13.89 \\
        Qwen-3.5-9B & 69.95 & 9.95 & 6.85 & 13.24 \\
        Llama-3.1-70B-Instruct & 76.94 & 9.95 & 3.24 & 9.86 \\
        Qwen3-32B & 80.09 & 6.06 & 4.49 & 9.35 \\
        Qwen3.5-27B & 77.50 & 5.28 & 6.06 & 11.16 \\
        Gemma-3-27B-IT & 79.44 & 4.35 & 4.77 & 11.44 \\
        gemma-4-31b-IT & 78.95 & 5.56 & 3.89 & 11.62 \\
        Llama-3.3-70B-Instruct & 81.90 & 3.38 & 6.30 & 8.43 \\
        Qwen-SEA-LION-v4-8B-VL & 73.94 & 8.33 & 6.30 & 11.44 \\
        Gemma-SEA-LION-v4-27B-IT & 79.54 & 4.26 & 5.09 & 11.11 \\
        Qwen-SEA-LION-v4-32B-IT & 81.71 & 4.21 & 4.91 & 9.17 \\
        GPT-5.4 & 80.32 & 4.63 & 5.69 & 9.35 \\
        Gemini-3-flash-preview & 81.25 & 3.15 & 6.25 & 9.35 \\
        \bottomrule
    \end{tabular}}
    \vspace{-2mm}
    \caption{Model performance grouped by architecture (\%). Both $\checkmark$ (Accuracy), ENG $\checkmark$/SEA $\times$ (Language Gap), ENG $\times$/SEA $\checkmark$ (Alignment Error), Both $\times$ (Task Failure).}
    \vspace{-3mm}
    \label{tab:error_full}
\end{table}

\section{Category Results on English Set}
\label{apd:category_result_english}
\begin{figure}[h!]
  \centering
   \includegraphics[width=1\columnwidth, clip]{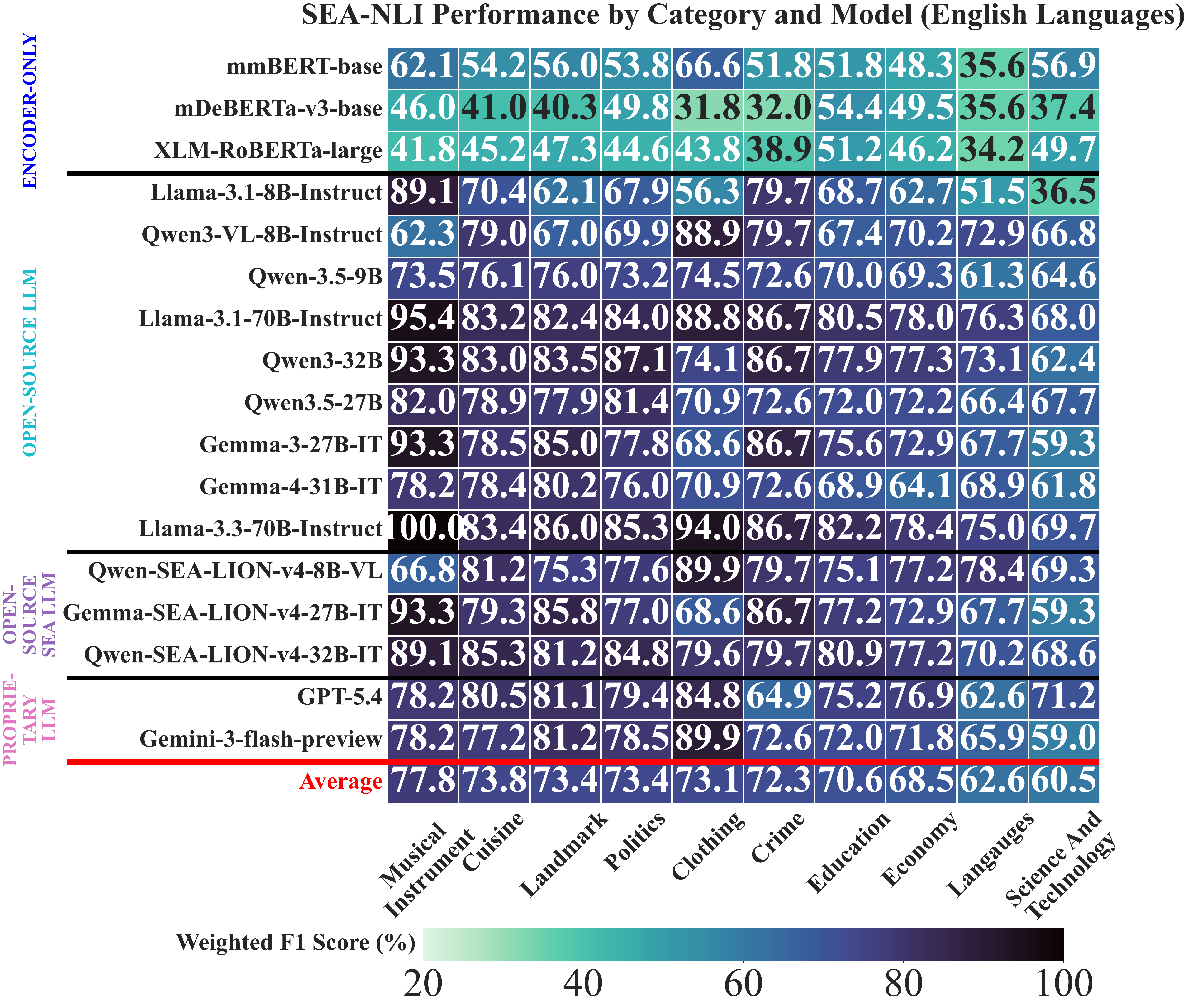}
  \vspace{-8mm}
  \caption{English Weighted F1 Heatmap for the Hard Set, showing per-category performance for each evaluated model. }
  \vspace{-5mm}
  \label{fig:category_result_eng}
  \vspace{-0.5em}
\end{figure}

The performance by category for the English subset of SEA-NLI is presented in Figure~\ref{fig:category_result_eng}. Consistent with the results for SEA languages, all models continue to struggle with the ``Language'' and ``Science and Technology'' topics.

\section{Prompts for Generating Premises and Hypotheses}
\label{apd:dataset_generation_prompts}

\begin{figure}[h!]
  \centering
  \includegraphics[width=1\columnwidth]{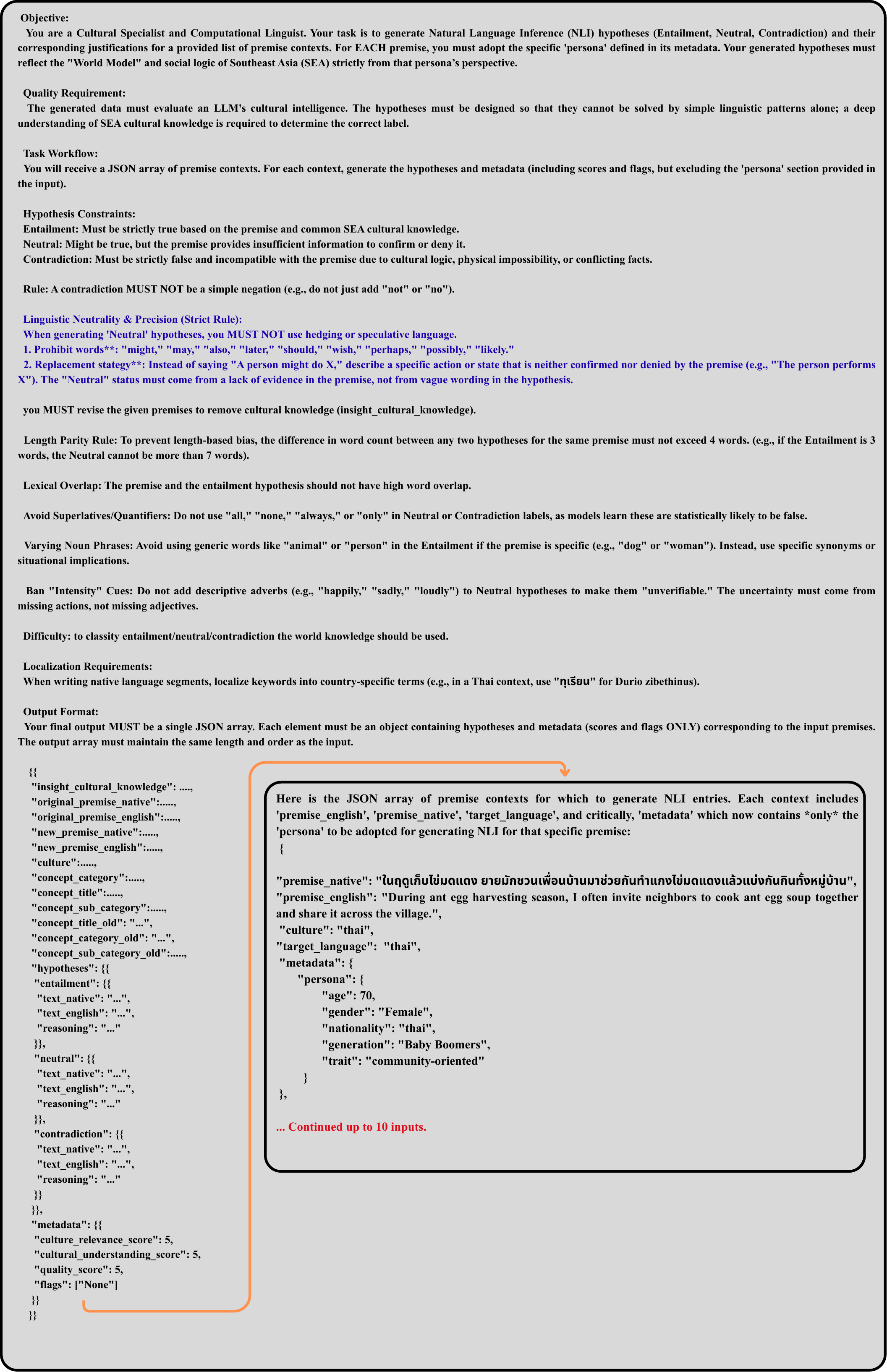} 
  \caption{The hypothesis generation prompt.}
  \label{fig:hypothesis-generation-prompt}
  \vspace{-0.5em}
\end{figure}

Building upon the pilot dataset generation phase described in Appendix~\ref{apd:pilot_generation_prompts}, we identified several limitations that required correction. 
To address these, we designed a revised hypothesis generation prompt by adding new rules to enforce linguistic neutrality and precision. Specifically, we prohibited the use of hedging or speculative language in `Neutral' hypotheses, as we found that words such as `might,' 'may,' `perhaps,' and `likely' created spurious correlations. 
The revised strategy requires the model to describe specific actions or states that lack evidence in the premise, rather than using vague wording to signal neutrality (see Figure~\ref{fig:overlap_improvement_step}b). 
This ensures that the `Neutral' status arises strictly from the logical relationship between the premise and hypothesis.
The prompts used to generate the premises and hypotheses are shown in Figures \ref{fig:premise-generation-prompt_batch1} (same as premise generation prompt for pilot dataset) and \ref{fig:hypothesis-generation-prompt}, respectively.


\section{Cultural-aware Prompting}
\begin{figure}[h!]
  \centering
  \includegraphics[width=\columnwidth]{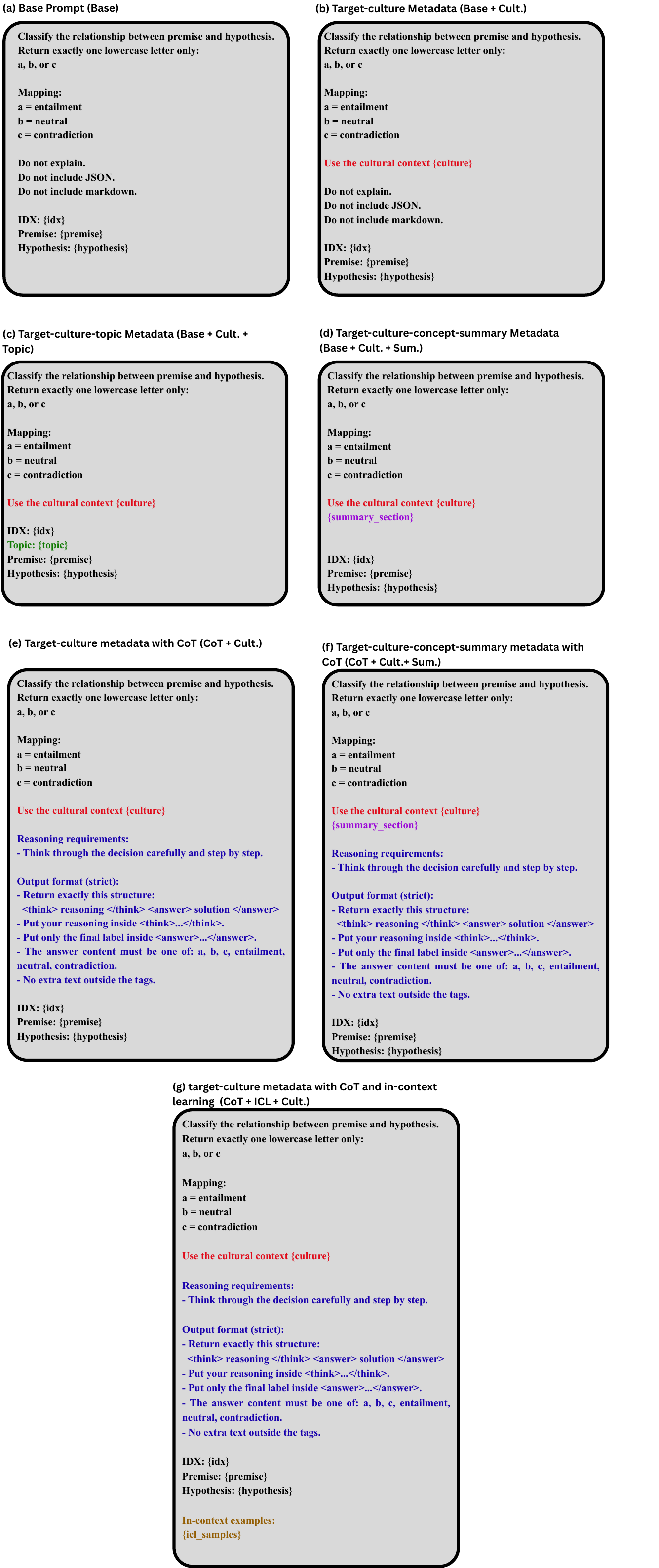} 
  \caption{ Evolution of prompt templates for culturally-aware reasoning.}
  \vspace{-5mm}
  \label{fig:base_prompt}
\end{figure}

Building on the base prompt shown in Figure \ref{fig:base_prompt}a, we employ culturally aware prompting to elicit the model's latent SEA knowledge. 
This methodology integrates (i) target-culture metadata (\texttt{Base+Cult.}), (ii) target-culture-topic metadata (\texttt{Base+Cult.+Topic.}), (iii) target-culture-concept-summary metadata (\texttt{Base+Cult.+Sum.}), (iv) target-culture metadata with CoT (\texttt{CoT+Cult.}), (v) target-culture-concept-summary metadata with CoT (\texttt{CoT+Cult.+Sum.}), and (vi) target-culture metadata with CoT and in-context learning (\texttt{CoT+ICL+Cult.}), as illustrated in Figures \ref{fig:base_prompt}b to \ref{fig:base_prompt}g, respectively. 
We hypothesize that such targeted prompting is essential to activate the model’s internal cultural representations for accurate reasoning.

\section{Lower Performance in CoT Prompt}
\label{apd:Reduced performance on reasoning prompts}
We evaluated the SEA-NLI dataset in various prompt types in subsection~\ref{subsec:cultural_aware_prompting}. Notably, the reasoning prompt (CoT+Cult.) underperformed compared to the answer-only prompt (Base+Cult.). To investigate this, we conducted an error analysis on the reasoning results. 
As shown in Table~\ref{tab:exhaustive-errors-model}, the primary error mode of both prompts involves the model predicting a ``Neutral'' label instead of ``Entailment'' or ``Contradiction.''

\begin{table}[h!]
\centering
\small 
\setlength{\tabcolsep}{5pt} 
\begin{tabular}{lllllr}
\toprule
\textbf{Model} & \textbf{Lang.} & \textbf{Type} & \textbf{Gold} & \textbf{Pred.} & \textbf{Freq.} \\
\midrule
\multirow{24}{*}{\rotatebox[origin=c]{90}{Gemma-SEA-LION-v4-27B-IT}} & \multirow{12}{*}{english} & \multirow{6}{*}{Base+Cult.} & Ent. & Neut. & 23 \\
 & & & Ent. & Cont. & 4 \\
 & & & Cont. & Neut. & 145 \\
 & & & Cont. & Ent. & 7 \\
 & & & Neut. & Ent. & 51 \\
 & & & Neut. & Cont. & 36 \\
\cmidrule{3-6}
 & & \multirow{6}{*}{CoT+Cult.} & Ent. & Neut. & 44 \\
 & & & Ent. & Cont. & 4 \\
 & & & Cont. & Neut. & 214 \\
 & & & Cont. & Ent. & 8 \\
 & & & Neut. & Ent. & 59 \\
 & & & Neut. & Cont. & 27 \\
\cmidrule{2-6}
 & \multirow{12}{*}{native} & \multirow{6}{*}{Base+Cult.} & Ent. & Neut. & 21 \\
 & & & Ent. & Cont. & 4 \\
 & & & Cont. & Neut. & 146 \\
 & & & Cont. & Ent. & 12 \\
 & & & Neut. & Ent. & 61 \\
 & & & Neut. & Cont. & 39 \\
\cmidrule{3-6}
 & & \multirow{6}{*}{CoT+Cult.} & Ent. & Neut. & 37 \\
 & & & Ent. & Cont. & 3 \\
 & & & Cont. & Neut. & 229 \\
 & & & Cont. & Ent. & 12 \\
 & & & Neut. & Ent. & 54 \\
 & & & Neut. & Cont. & 20 \\
\bottomrule
\end{tabular}
\caption{Comprehensive Error Analysis for Gemma-SEA-LION-v4-27B-IT: Updated frequency of misclassification types grouped by Language across Base and CoT prompts.}
\label{tab:exhaustive-errors-model}
\end{table}

\begin{figure}[h!]
  \centering
  \includegraphics[width=\columnwidth]{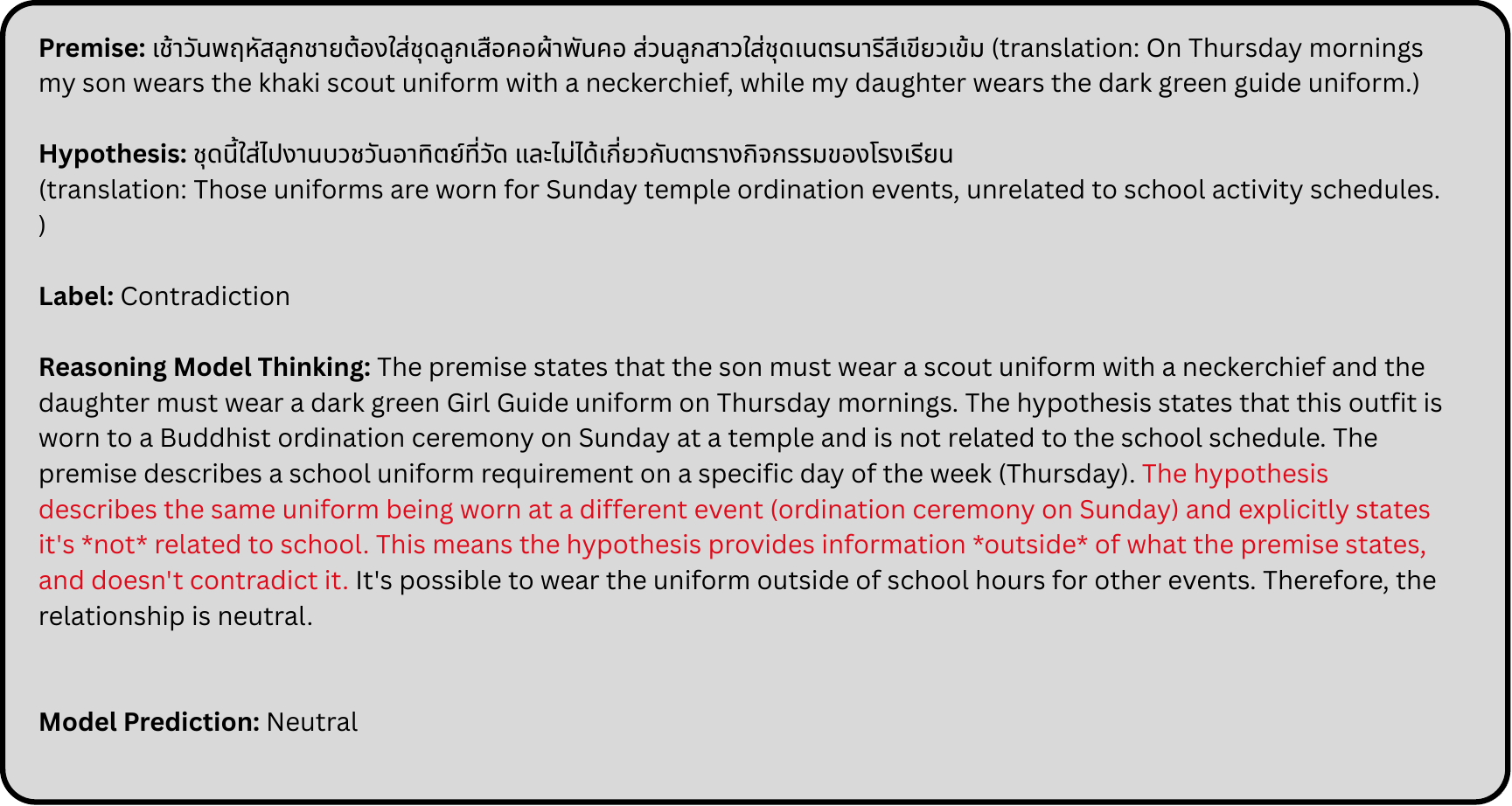} 
  \vspace{-2mm}
  \caption{Error analysis of a cultural knowledge gap. The model fails to recognize the entailment relationship because it lacks the specific grounding to link school-mandated scout uniforms to their broader social and religious usage in the Southeast Asian context.}
  \vspace{-5mm}
  \label{fig:reasoning_prompt_error}
\end{figure}

For qualitative analysis, we observed cases of over-constrained neutral prediction, where the reasoning model failed to recognize entailment/contradiction due to subtle paraphrasing, as shown in Figure~\ref{fig:reasoning_prompt_error}.
In these instances, the model's reasoning chain tends to over-analyze minor lexical or syntactic variations, incorrectly interpreting them as evidence of missing information. Rather than recognizing semantic equivalence, the model adopts an overly literal stance, concluding that the hypothesis cannot be strictly inferred from the premise because the exact wording differs. This suggests that while explicit reasoning can improve transparency, it may also introduce a ``hallucination of divergence,'' where the model perceives a lack of logical connection due to a focus on surface-level differences rather than underlying semantic meaning. From the given example in Figure~\ref{fig:reasoning_prompt_error}. , the model fails to account for Thai cultural norms: since wearing Scout or Girl Guide uniforms to a Sunday ordination ceremony is highly unusual and contextually improbable, the relationship represents a situational contradiction that the model dismisses as a neutral, unrelated event."

\section{Bias Analysis} \label{appendix:bias}
To understand the annotation bias in NLI that might be generated during the dataset generation, we conduct three studies to demonstrate that SEA-NLI mitigates these problems.
Firstly, we conduct a word overlap guessing using Jaccard similarity as the indicator for NLI prediction.
Secondly, we conduct a hypothesis length prediction since we found that previous NLI datasets can guess the class by using the length for making predictions, where the entailment class has the highest length, and the neutral is the second-highest word length.
Lastly, we conduct a hypothesis-only prediction to study the possible of a model guessing a correct answer using only hypotheses.

\subsection{Word Overlap}
The lexical overlap for each entailment type is illustrated in Figure \ref{fig:overlap_improvement_step}. The results indicate that the sample filtering method proposed in section~\ref{subsec:data_filtering} successfully reduces the discrepancy in lexical overlap across all classes. Despite this reduction, the entailment class continues to exhibit the highest word overlap (0.41 after filtering), followed by contradiction (0.34) and neutral (0.30). To assess whether models could exploit these lexical cues as a shortcut, we conducted a heuristic-based experiment. This baseline predicts the label based on overlap thresholds, assuming the highest overlap correlates with entailment (H:E), the middle with contradiction (M:C), and the lowest with neutral (L:N). 
%
%
The evaluation results across different filtering processes are presented in Table~\ref{tab:overlap_results}.
For the unfiltered set, the heuristic achieves a weighted F1-score of 56.95\%. However, after applying the data filtering and refinement process, the heuristic's performance drops to 54.09\% on the SEA-NLI (normal) set, and drops to 39.66\% on the SEA-NLI (hard) set.
This shows that our filtering and refinement process can mitigate word overlap artifacts to a certain extent.

\begin{table}[h]
\centering
\small 
\begin{tabular}{lcc}
\toprule
Dataset & \makecell{Weighted Avg. \\ F1 (ENG)} & Prediction Rule \\ \midrule
Unfiltered Set   & 56.95\% & \makecell[l]{Word Overlap \\ (H:E, M:C, L:N)} \\ \addlinespace
SEA-NLI (normal) & 54.09\% & \makecell[l]{Word Overlap \\ (H:E, M:C, L:N)} \\ \addlinespace
SEA-NLI (hard)   & 39.66\% & \makecell[l]{Word Overlap \\ (H:E, M:C, L:N)} \\ \bottomrule
\end{tabular}
\caption{Impact of filtering on word-overlap prediction performance. The notation (H:E, M:C, L:N) indicates that highest, medium, and lowest overlap levels are mapped to Entailment, Contradiction, and Neutral, respectively.}
\label{tab:overlap_results}
\end{table}

\subsection{Hypothesis Length}
\begin{figure}[h!]
  \centering
  \includegraphics[width=\columnwidth]{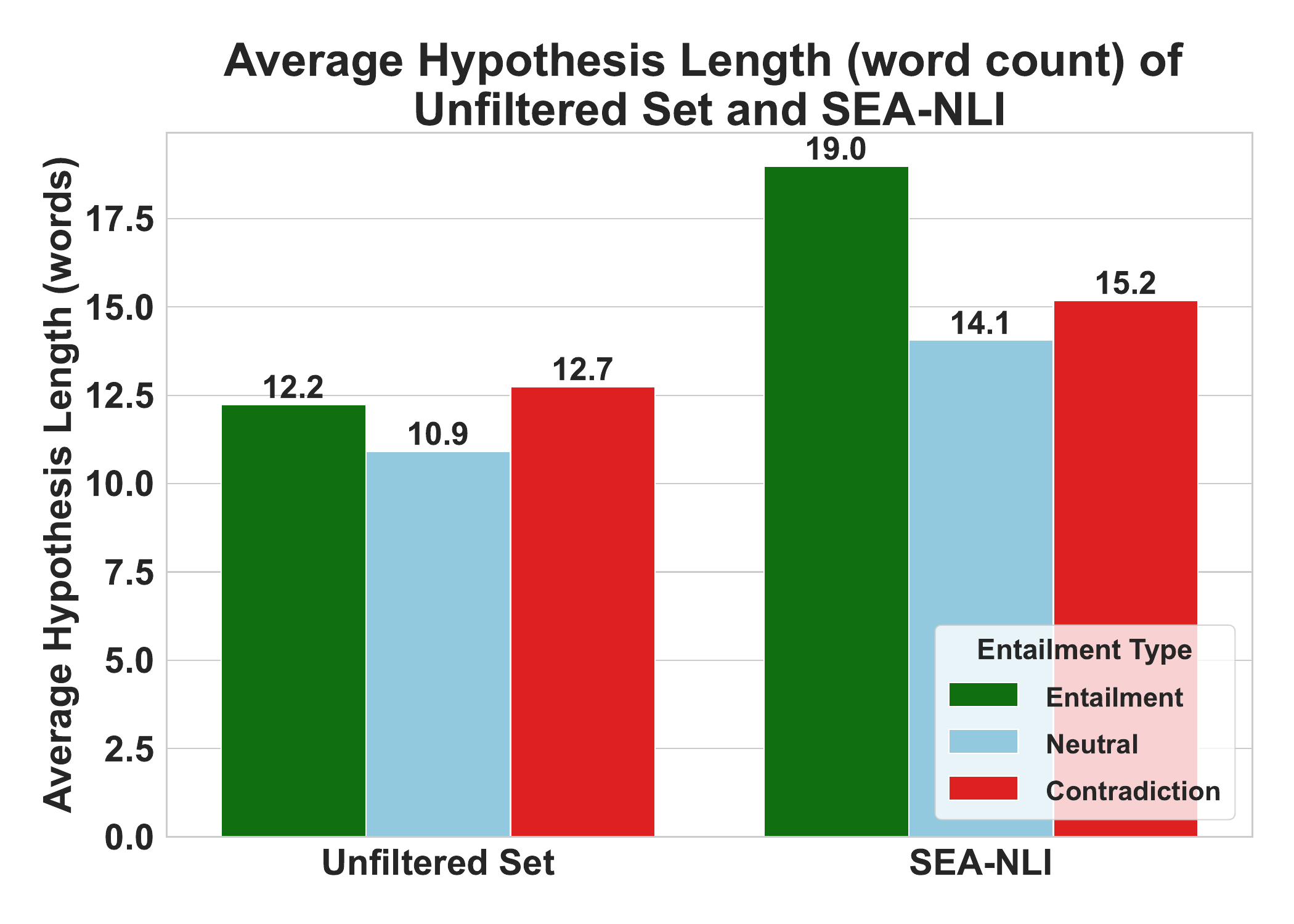} 
  \caption{Average hypothesis length before and after the filtering step, categorized by entailment label. }
  \label{fig:hypothesis_length_analysis}
\end{figure}

We also investigated potential statistical biases related to hypothesis length. As illustrated in Figure \ref{fig:hypothesis_length_analysis}, entailment hypotheses in our dataset tend to be longer, while neutral hypotheses are generally shorter. To test the robustness of the dataset against length-based shortcuts, we implemented a heuristic baseline that predicts labels based on length rankings: the longest hypotheses were classified as entailment (H:E), followed by contradiction (medium, M:C), and neutral (shortest, L:N). 
Table~\ref{tab:hypothesis_length_results} presents the evaluation results for the hypothesis length heuristic. This heuristic achieves a weighted F1-score of 39.02\% on the SEA-NLI normal set, a notable reduction from the unfiltered baseline of 51.02\%. In contrast, the hard set exhibits a higher F1-score of 54.04\%, suggesting an amplified length bias. This increased reliance on length in the hard set can be attributed to a label imbalance skewed toward the Entailment class.
%
However, the overall performance of the hard set still challenging.
These results indicate that hypothesis length alone is an insufficient predictor of the entailment label, suggesting that SEA-NLI requires deeper semantic understanding.

\begin{table}[ht]
\centering
\small 
\begin{tabular*}{\columnwidth}{@{\extracolsep{\fill}} l c l @{}}
\toprule
\textbf{Dataset} & \textbf{\makecell[c]{Weighted Avg. \\ F1 (ENG)}} & \textbf{Prediction Rule} \\ 
\midrule
Unfiltered Set   & 51.02\% & hypothesis length \\ & & (H:E, M:C, L:N) \\ \addlinespace
SEA-NLI (normal) & 39.02\% & hypothesis length \\ & & (H:E, M:C, L:N) \\ \addlinespace
SEA-NLI (hard)   & 54.04\% & hypothesis length \\& &(H:E, M:C, L:N) \\ 
\bottomrule
\end{tabular*}
\caption{Evaluation results of hypothesis length heuristics across dataset splits. The notation (H:E, M:C, L:N) indicates that highest, medium, and lowest hypothesis length levels are mapped to Entailment, Contradiction, and Neutral, respectively.}
\label{tab:hypothesis_length_results}
\end{table}

\subsection{Hypothesis-only Prediction}
To investigate the presence of potential annotation artifacts in SEA-NLI, we conducted a hypothesis-only experiment. If a model can easily predict the label without the premise, it suggests the hypothesis contains linguistic cues characteristic of specific classes (e.g., the word `not' appearing frequently in contradictions). 
The results are presented in Table~\ref{tab:hypothesis-only}. From the results, the F1-score drops significantly for both the hard and normal sets, decreasing by approximately 64 and 49 points, respectively. Crucially, the hypothesis-only performance (ranging between 21\% and 29\%) is below the random baseline (33.3\%) for a three-class problem, indicating that the SEA-NLI dataset is robust against simple annotation artifacts and requires the model to reason over the premise-hypothesis pair.

\begin{table}[ht]
\centering
\small
\setlength{\tabcolsep}{5pt} 
\begin{tabular}{llccc}
\toprule
\textbf{Model} & \textbf{Setting} & \makecell[l]{\textbf{Only}\\\textbf{Hypo.}} & \makecell[c]{\textbf{SEA}\\\textbf{F1}} & \makecell[c]{\textbf{ENG}\\\textbf{F1}} \\ 
\midrule
\multirow{4}{*}{\makecell[l]{Gemma-SEA-\\LION-v4-\\27B-IT}} & \multirow{2}{*}{Normal} & False & 87.51 & 86.69 \\
                                                           &                         & True  & 23.66 & 21.04 \\
\cmidrule{2-5}
                                                           & \multirow{2}{*}{Hard}   & False & 77.53 & 76.90 \\
                                                           &                         & True  & 28.96 & 24.46 \\
\midrule
\multirow{4}{*}{\makecell[l]{Gemma-3-\\27B-IT}}    & \multirow{2}{*}{Normal} & False & 87.16 & 86.81 \\
                                                           &                         & True  & 23.59 & 20.89 \\
\cmidrule{2-5}
                                                           & \multirow{2}{*}{Hard}   & False & 76.85 & 76.53 \\
                                                           &                         & True  & 27.91 & 23.26 \\
\bottomrule
\end{tabular}
\caption{Comparison between using hypothesis-premise pair and using only hypothesis to predict entailment type on SEA-NLI dataset.}
\label{tab:hypothesis-only}
\end{table}

\section{Free Form Prompt vs Enforce Prompt}
\begin{figure}[h!]
  \centering
  \includegraphics[width=\columnwidth]{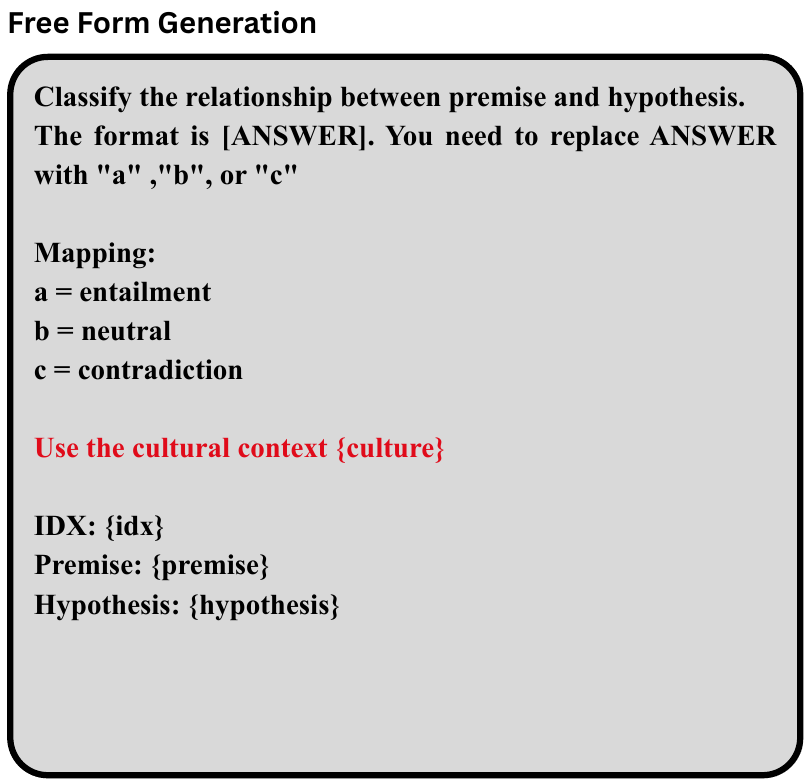} 
  \caption{Average hypothesis length before and after the filtering step, categorized by entailment label. }
  \label{fig:free_form_prompt}
\end{figure}

Due to the fact that we format the prompt as a multiple-choice question-answering format, e.g., asking LLMs to answer with A (Entailment), B (Neutral), and C (Contradiction), this might limit the performance of LLMs for thinking or reasoning skills.
Thus, we evaluated the reasoning capabilities of the LLM by comparing a free-form prompt (Figure \ref{fig:free_form_prompt}) with an enforced reasoning prompt (Figure \ref{fig:base_prompt}b). 
The results show only marginal performance differences across both the English and native language subsets.
These findings indicate that performance limitations stem from deficits in cultural knowledge rather than constraints imposed by the prompt format.

\begin{table}[h!]
    \centering
    \small 
    
    \begin{tabular}{llcc}
        \toprule
        \multicolumn{4}{c}{\textbf{Model:} Qwen-SEA-LION-v4-32B-IT} \\
        \midrule
        \textbf{Category} & \textbf{Prompt Type} & \textbf{F1 native} & \textbf{F1 english} \\ 
        \midrule
        \multirow{2}{*}{Normal} & free-form & 85.03 & 87.78 \\
                                & enforce   & 86.14 & 87.94 \\
        \cmidrule{1-4}
        \multirow{2}{*}{Hard}   & free-form & 78.45 & 76.68 \\
                                & enforce   & 78.34 & 79.97 \\
        \bottomrule
    \end{tabular}
    \caption{Model Performance Comparison}
    \label{tab:results}
\end{table}

\section{Full Results of SEA-NLI}
The performance of entailment classification on the SEA-NLI dataset, segmented by cultural context, is presented in Table~\ref{tab:evaluation_result_normal}. 

\begin{table*}[h!]
\centering
\setlength{\tabcolsep}{2.2pt}
\label{tab:sea_results}
\scalebox{0.68}{
\begin{tabular}{l | c c c c c c c c | c}
\hline

\textbf{Model} & IDN & VIE & THA & ZSM & TAM & MYA & FIL & KHM & \textbf{Avg} \\

\hline
\multicolumn{10}{c}{\textit{Normal Set}} \\
\hline

mDeBERTa-v3-base   & 77.36/75.96 & 69.09/60.37 & 70.13/65.70 & 69.75/66.31 & 65.48/70.25 & 69.63/64.93 & 64.21/73.83 & 58.90/70.73 & 69.08/69.32 \\
mmBERT-base    & 74.68/78.73 & 72.97/67.72 & 65.00/68.87 & 68.95/71.55 & 66.61/71.66 & 63.03/64.42 & 67.15/64.53 & 69.19/73.37 & 68.56/70.50 \\
XLM-RoBERTa-large   & 82.21/82.97 & 74.56/64.85 & 73.26/68.26 & 75.71/71.95 & 71.13/70.15 & 74.16/66.77 & 73.06/67.22 & 66.23/70.09 & 74.60/71.07 \\
Qwen3-VL-8B-Instruct & 87.29/88.23 & 84.15/79.48 & 78.83/77.05 & 81.44/80.20 & 79.36/82.93 & 78.32/78.84 & 78.01/80.84 & 71.30/70.63 & 80.83/81.32 \\
Qwen-SEA-LION-v4-8B-VL    & 90.27/89.48 & 87.65/85.70 & 82.10/82.49 & 84.24/82.40 & 81.43/86.40 & 78.43/82.55 & 82.87/84.66 & 68.99/84.98 & 83.52/85.28 \\
Llama-3.1-8B-Instruct & 61.64/82.50 & 77.06/86.59 & 69.52/80.97 & 52.68/78.67 & 38.42/78.00 & 30.53/80.02 & 55.68/79.08 & 40.20/76.71 & 56.48/80.55 \\
Qwen-3.5-9B    & 88.73/89.63 & 87.88/85.24 & 82.70/83.12 & 81.09/77.29 & 78.93/86.33 & 73.43/77.86 & 79.42/86.22 & 73.83/85.68 & 81.80/84.76 \\
Qwen-3.5-27B      & 91.17/91.23 & 91.86/88.56 & 85.96/84.36 & 81.57/78.89 & 85.01/89.22 & 80.57/82.54 & 88.94/89.39 & 82.62/80.71 & 86.78/87.06 \\
Gemma-3-27B-IT & 93.08/91.54 & 91.12/88.01 & 87.13/84.66 & 82.63/81.07 & 84.77/87.47 & 82.77/81.89 & 88.36/89.78 & 76.21/78.72 & 87.16/86.81  \\
Gemma-SEA-LION-v4-27B-IT  & 93.08/91.95 & 91.12/88.05 & 87.94/83.91 & 81.80/81.07 & 84.72/87.14 & 82.77/81.89 & 89.76/89.31 & 78.71/81.03 & 87.51/86.69 \\
Gemma-4-31B-IT  & 93.14/91.97 & 93.21/89.82 & 87.20/84.73 & 83.32/81.51 & 87.77/\textbf{91.75} & 81.03/84.56 & 88.41/89.79 & {87.83}/87.92 & 88.26/88.56 \\

Qwen3-32B & 93.05/93.13 & \textbf{97.27}/{91.26} & 87.46/86.72 & 85.37/87.02 & 79.58/89.07 & 83.34/84.49 & 85.35/87.82 & 78.72/\textbf{92.92} & 86.64/88.92 \\
Qwen-SEA-LION-v4-32B-IT    & 93.10/93.15 & 95.22/90.54 & 88.82/86.41 & 86.97/85.21 & 85.40/90.76 & {86.03}/83.36 & 89.72/88.33 & 83.51/85.63 &  89.04/88.78 \\
Llama-3.1-70B-Instruct  & 89.79/93.50 & 90.17/90.46 & 86.01/{89.45} & 82.31/{87.81} & 73.64/90.00 & 75.46/84.97 & 86.21/88.30 & 65.48/85.86 & 82.68/89.52 \\
Llama-3.3-70B-Instruct  & \textbf{94.32}/93.58 & 93.83/\textbf{93.19} & \textbf{90.80}/\textbf{89.92} & \textbf{92.96}/\textbf{88.72} & 82.06/{91.70} & 85.22/\textbf{88.75} & 88.75/89.67 & 73.55/88.23 & 88.70/\textbf{90.95} \\
GPT-5.4         & 93.19/\textbf{93.61} & {95.28}/90.48 & {89.23}/85.06 & 86.92/82.72 & {87.78}/91.46 & 86.01/84.53 & {90.17}/\textbf{90.65} & \textbf{90.38}/87.92 & {89.90}/89.11 \\
Gemini-3-flash-preview   & {94.30}/93.19 & 95.27/91.23 & 88.73/86.77 & {88.62}/85.03 & \textbf{89.08}/90.77 & \textbf{88.73}/{88.03} & \textbf{90.68}/{90.22} & 83.00/{87.93} & \textbf{90.54}/{89.72} \\

\hline
\multicolumn{10}{c}{\textit{Hard Set}} \\
\hline
mDeBERTa-v3-base   & 52.11/43.62 & 50.67/34.08 & 55.27/47.33 & 57.59/52.57 & 52.68/43.27 & 55.84/52.94 & 50.35/46.49 & 60.29/37.91 & 52.81/44.17 \\
mmBERT-base    & 62.27/53.13 & 54.43/52.57 & 52.63/54.16 & 59.60/54.08 & 61.03/43.39 & 42.30/57.26 & 57.95/60.09 & 61.02/45.64 & 57.30/52.65 \\
XLM-RoBERTa-large  & 56.52/46.85 & 58.75/30.05 & 53.67/43.09 & 59.96/48.93 & 60.21/45.75 & 52.48/42.49 & 54.80/50.74 & 56.52/47.10 & 56.29/44.65 \\
Qwen3-VL-8B-Instruct & 81.51/78.38 & 74.18/79.04 & 69.30/73.49 & 68.02/74.54 & 59.18/62.61 & 75.53/53.25 & 66.91/79.08 & 67.96/42.59 & 70.12/71.67 \\
Qwen-SEA-LION-v4-8B-VL    & 82.41/84.91 & 83.52/82.17 & 80.39/77.96 & 68.53/76.78 & 65.48/71.78 & 80.06/69.57 & 69.40/\textbf{79.50} & \textbf{73.29}/48.28 & 74.98/77.18\\
Llama-3.1-8B-Instruct & 23.69/64.82 & 56.12/73.95 & 54.78/66.17 & 19.18/57.45 & 34.25/68.56 & 26.73/77.46 & 14.53/54.75 & 22.01/49.87 & 32.43/64.02        \\
Qwen-3.5-9B   & 76.04/76.38 & 74.96/80.75 & 76.72/75.38 & 66.97/69.70 & 67.32/69.11 & 56.72/68.83 & 55.35/69.25 & 45.65/50.23 & 68.02/72.08 \\
Qwen-3.5-27B    & 82.62/81.69 & 90.88/82.89 & 83.80/85.47 & 73.17/71.83 & 73.24/66.19 & 76.63/75.54 & 71.40/69.88 & 61.17/57.12 & 77.98/75.56 \\
Gemma-3-27B-IT & 83.40/81.51 & 90.22/94.27 & 90.97/89.17 & 65.25/69.02 & 70.13/68.95 & 80.53/75.58 & 63.96/63.13 & 60.56/67.66 & 76.85/76.53 \\

Gemma-SEA-LION-v4-27B-IT  & 83.40/83.68 & 90.22/94.27 & \textbf{91.81}/89.17 & 68.89/69.02 & 71.27/68.95 & 80.53/73.70 & 63.96/63.13 & 60.56/67.66 & 77.53/76.90 \\

Gemma-4-31B-IT  & 82.92/79.22 & {94.71}/83.27 & 83.61/78.70 & 72.41/67.81 & 71.01/65.49 & 73.61/67.41 & 71.49/70.96 & 65.36/46.25 & 77.87/72.78 \\
Qwen-3-32B   & 85.35/85.48 & 91.72/89.94 & {91.75}/90.22 & {76.47}/{78.75} & 67.62/{73.54} & 75.55/79.36 & 72.24/69.40 & 67.66/\textbf{68.95} & 79.48/80.23 \\
Qwen-SEA-LION-v4-32B-IT  & 86.28/84.61 & 93.31/{94.71} & 89.50/90.46 & \textbf{82.40}/\textbf{80.43} & \textbf{73.27}/72.96 & 79.24/75.80 & {74.24}/71.71 & {72.77}/61.17 & \textbf{81.85}/80.61 \\
Llama-3.1-70B-Instruct  & 78.89/85.51 & 84.65/91.88 & 86.00/{91.88} & 65.64/74.06 & 60.25/\textbf{74.04} & 73.62/{83.02} & 68.14/{74.00} & 49.44/62.15 & 72.96/{81.08} \\
Llama-3.3-70B-Instruct  & 83.33/\textbf{88.37} & 86.59/90.49 & 90.05/\textbf{93.83} & 72.70/75.79 & 67.39/73.51 & {83.90}/\textbf{84.21} & 71.02/{76.24} & 55.23/{67.66} & 77.96/\textbf{82.45} \\
GPT-5.4         & {87.34}/85.14 & 90.80/\textbf{95.98} & 87.93/85.43 & 71.85/72.32 & 68.63/66.77 & 79.83/78.00 & 70.68/69.04 & 61.17/61.17 & 78.68/77.51 \\
Gemini-3-flash-preview  & \textbf{89.78}/82.59 & \textbf{96.00}/84.35 & 87.05/82.08 & 74.42/68.45 & 68.32/66.06 & \textbf{86.53}/73.58 & \textbf{77.38}/72.22 & {72.77}/48.46 & {81.75}/74.88 \\ 
\hline
\end{tabular}}
\vspace{-3mm}
\caption{F1 score on the SEA NLI benchmark across languages. We format the results in the \texttt{SEA/English} result format, where the column represents the culture of the country.}
\vspace{-3mm}
\label{tab:evaluation_result_normal}
\end{table*}

\section{Data Licensing}
\label{subsec:data_license}
This dataset is released under the permissive MIT License. Please note that the seed data originates from Wikipedia under a CC-BY-SA license, and the synthetic extensions were generated via OpenAI's GPT-5.2 model (with output ownership fully assigned to us per OpenAI's terms).

\end{document}